18
\documentclass[runningheads]{llncs}
\usepackage{graphicx}
\usepackage{booktabs}
\usepackage{float}

\restylefloat{table}
\usepackage{array}
\usepackage{geometry}
\usepackage{tabularx}
\usepackage{adjustbox}
\usepackage[T1]{fontenc}
\UseRawInputEncoding

\usepackage{multirow}

\begin{document}

\title{A survey on cutting-edge relation extraction techniques based on language models}
%
%
\author{J. Angel Diaz-Garcia \orcidID{0000-0002-9263-1402} \and Julio Amador Diaz Lopez }
\authorrunning{Diaz-Garcia and Amador Diaz Lopez }
\titlerunning{A survey on cutting-edge relation extraction}
%
\institute
{
    Department of Computer Science and A.I., University of Granada, Spain
     \email{jagarcia@decsai.ugr.es} \\
    SiftyML, UK
    \email{julio@siftyml.com}
}

\maketitle         

%

%
\begin{abstract}
This comprehensive survey delves into the latest advancements in Relation Extraction (RE), a pivotal task in natural language processing essential for applications across biomedical, financial, and legal sectors. This study highlights the evolution and current state of RE techniques by analyzing 137 papers presented at the Association for Computational Linguistics (ACL) conferences over the past four years, focusing on models that leverage language models. Our findings underscore the dominance of BERT-based methods in achieving state-of-the-art results for RE while also noting the promising capabilities of emerging large language models (LLMs) like T5, especially in few-shot relation extraction scenarios where they excel in identifying previously unseen relations.

\end{abstract}
\section{Introduction}

With the rapid growth of data generated and stored on the internet, numerous tools have emerged to process and extract valuable insights from vast information. Given that a significant portion of this data exists as unstructured text, the need for techniques capable of handling such data is evident. The branch of artificial intelligence dedicated to processing text is Natural Language Processing (NLP) \cite{ranjan2016survey}. NLP provides practitioners and researchers with a comprehensive toolkit for effectively analyzing unstructured text from diverse sources, including social media \cite{diaz2024flexible,coppersmith2018natural}, public health data \cite{hossain2023natural}, research papers \cite{de2024can} and digital humanities \cite{suissa2022text}. 

One of the prominent topics within NLP is Relation Extraction (RE). RE is a task focusing on identifying and extracting the intricate relationships between different entities mentioned in the textual content. Essentially, RE automates discovering connections between words or phrases within a text. For example, in the sentence ``The Eiffel Tower is located in Paris,'' a relation extraction system would automatically identify the relationship "located in" between the entities ``Eiffel Tower'' and ``Paris.'' Though this example may seem trivial, it can be extrapolated to other domains with far-reaching implications. In fields such as biomedicine, finance, and law \cite{thomas2022adaptable}, the importance of Relation Extraction becomes undeniable, as it enables the automatic discovery of critical connections within vast amounts of unstructured data.\footnote{
In the biomedical domain for example, Relation Extraction (RE) plays a crucial role in identifying the complex interactions between compounds and proteins. This is particularly significant when analyzing signal transduction mechanisms and biochemical pathways. Protein-protein interactions, for instance, can initiate a wide range of biological processes. RE automates the extraction of these interactions and relationships from vast volumes of unstructured biomedical literature. This not only accelerates the research process but also enhances the precision of data extraction, enabling researchers to systematically uncover critical insights that would be difficult to identify manually.}

Relation Extraction (RE) is not just essential for identifying relationships within text, but also indispensable for downstream tasks such as question-answering systems, decision support systems, and expert systems. Its utility extends to the humanities, where it enhances knowledge acquisition in graph-based datasets, enabling more sophisticated and accurate information retrieval and analysis.

In recent years, the advancement of RE has been significantly bolstered by the integration of language models. These models, particularly those based on deep learning architectures, have demonstrated a remarkable ability to capture the semantic nuances of textual data. Language models enhance RE by enabling more accurate identification of relationships between entities by effectively understanding and representing the underlying meaning of words and phrases. This capability allows RE systems to move beyond surface-level associations, capturing complex and context-dependent relationships, thereby improving the precision and applicability of RE in various domains.

In this review, we aim to consolidate the latest advancements in RE, particularly emphasizing how recent research has leveraged language models. We aim to provide a comprehensive toolkit encompassing tasks, techniques, models, and datasets. By doing so, we intend to offer a resource that will serve as a springboard for future research and development.

Our survey overviews cutting-edge RE methodologies showcased at the Association for Computational Linguistics (ACL) conferences over the past three years. Numerous surveys on relation extraction have been conducted in the literature, including the notable work by Bassignana and Plank \cite{bassignana-plank-2022-mean}. In their paper, authors delve into the definition of relation extraction, categorizing techniques based on their subparts, such as named entity recognition, relation identification, and relation classification. Notably, their survey emphasizes scientific relation extraction, exclusively considering papers from ACL conferences from 2017 to 2021. We extend the analysis to satellite conferences, NAACL, AACL, and EACL. Our focus lies in utilizing language models, and we extend the review timeline to 2023, providing an updated perspective on trends in RE research. It is noteworthy to highlight the study conducted by Wadhwa et al. \cite{wadhwa-etal-2023-revisiting}, which provides a comprehensive review of the impact of Large Language Models (LLMs) like GPT or T5 across various benchmark datasets and baselines. While our work closely relates to theirs, our objective is to expand the scope of analysis beyond LLMs. We aim to incorporate a broader range of language models, conduct thorough comparisons, and undertake an extensive and robust review, encompassing a more comprehensive understanding of the field.

Our review aims to answer the following research questions (RQ): 

 \begin{itemize}

\item \textbf{RQ1}: What are the challenges of RE that are being solved by systems that leverage language models?

\item \textbf{RQ2}: What are the most commonly used language models for the RE problem?

\item \textbf{RQ3}: Which datasets are used as benchmarks for RE using language models?

\item \textbf{RQ4}: Are new large language models like GPT or T5 useful in RE versus widespread models such as BERT? 

 \end{itemize}

The subsequent sections of the paper are organized as follows: Section \ref{sec:methodology} delves into the intricate details of the methods employed in crafting this comprehensive review. Providing a foundational understanding of essential concepts, Section \ref{sec:background} offers a concise overview. Section \ref{sec:datasets} provides a comprehensive analysis of the datasets used for RE to the best of our knowledge. Our examination of cutting-edge RE models unfolds in Section \ref{sec:review}, while Section \ref{sec:llmvslm} scrutinizes the performance disparities between Language Models (LMs) and LLMs. Addressing the research questions that guided our exploration, Section \ref{sec:discusion} articulates responses derived from our review and findings. Finally, in Section \ref{sec:conclusion}, we briefly summarize the key takeaways and insights presented in this paper.

\section{Methodology}
\label{sec:methodology}

To ensure that we focus on cutting-edge applications of RE in natural language processing, we have limited our survey to the most recent papers presented at the Association for Computational Linguistics (ACL) conferences. These conferences are widely recognized as one of the most important and prestigious conferences in the field of NLP. Specifically, we focus on  ACL, NAACL, AACL, and EACL. This review was conducted from September 2023 to January 2024. During this period, the most recent conferences were ACL 2023, AACL 2023, and EACL 2023, marking the endpoint of our survey. We started in 2020, the first year with the significant incorporation of language models like BERT, which was introduced at NAACL 2019. It is important to note that ACL is the most prominent conference in our survey, as AACL, NAACL, and EACL are held biennially.

\begin{figure}[hbt]
  \centering
  \includegraphics[width=0.9\linewidth]{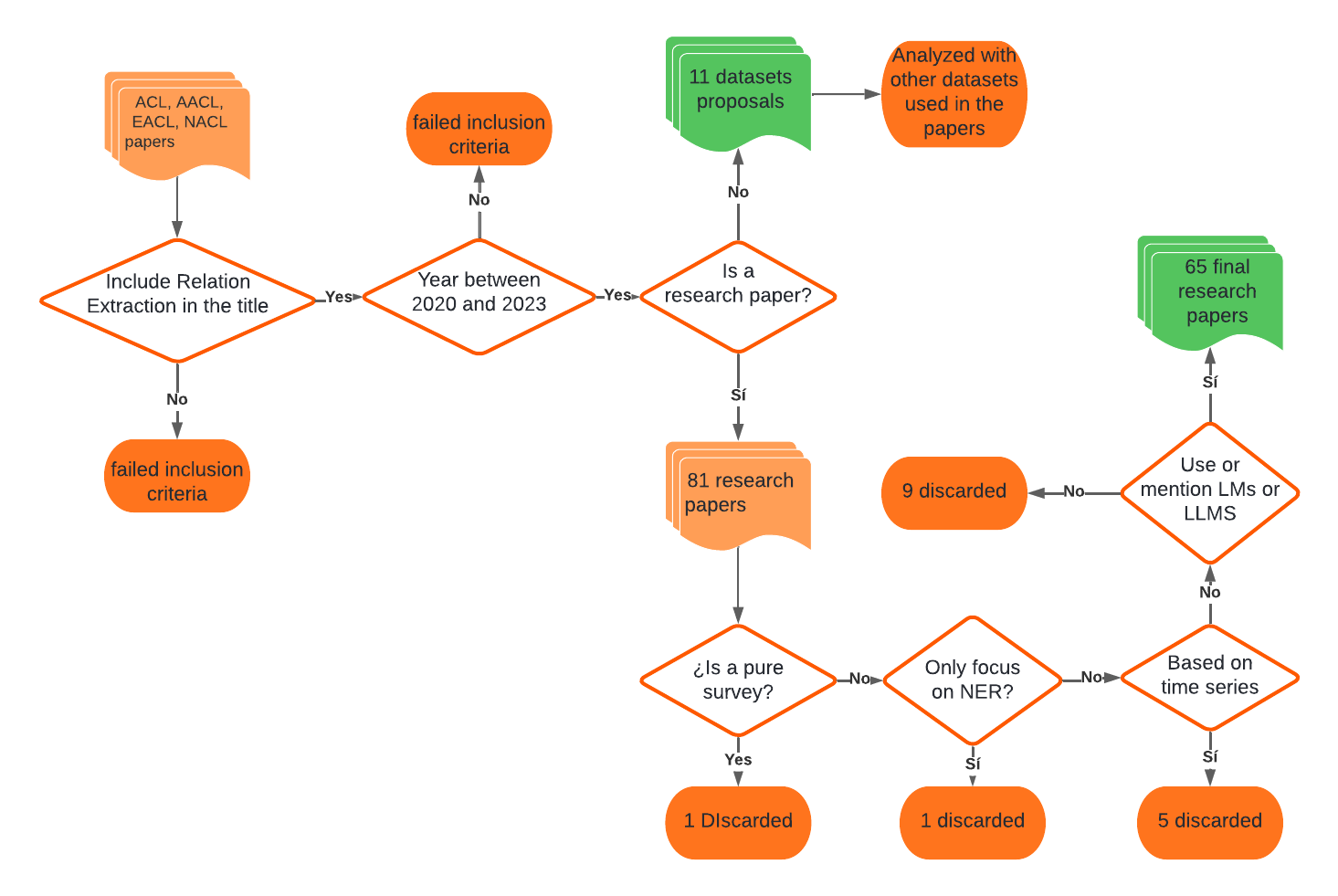}
  \caption{Graphic explanation of our inclusion/exclusion criteria}
  \label{fig:surveyflow}
\end{figure}

To narrow down the papers, we searched for pieces that contained the phrase ``Relation Extraction" in either the title or abstract. This search resulted in a set of papers screened based on their relevance to our research question and quality. Specifically, we only included papers that presented novel approaches or significant advancements in RE techniques using state-of-the-art language models. We exclude papers that are not directly related to the traditional problem of RE and venture into novel domains, such as temporal RE \cite{mathur2021timers,cui-etal-2021-refining,tan-etal-2023-event} or papers with a particular focus on NER \cite{ye-etal-2022-packed}. We also omitted papers that lack a foundation in language models or, at the very least, word embeddings \cite{zhang-etal-2021-open,yu-etal-2022-relation}. Our final exclusion criterion involves papers suggesting novel encoding techniques that, while intriguing, lack widespread adoption within the research community. For instance, the approach presented by \cite{vakil-amiri-2022-generic}, which introduces an exciting graph encoding solution, falls into this category. According to our inclusion criteria, the final set comprised 65 papers, classified as research contributions due to their introduction of new models, training strategies, or innovative approaches. For a better understanding, Figure \ref{fig:surveyflow} illustrates the complete process of our inclusion and exclusion criteria.

In total, we examined 81 papers presented at the last four years' editions of ACL conferences. Following inclusion criteria, the final set comprised 65 papers, classified as research contributions due to their introduction of new models, training strategies, or innovative approaches. Our investigation delved into the language models featured in these papers, with prominent examples including BERT and RoBERTa. Additionally, we scrutinized the diverse range of datasets utilized for model evaluation, encompassing well-known benchmarks like TACRED, NYT10, and DocRED. We examined a total of 56 datasets in our study. It is important to note that, in these instances, datasets are not exclusively confined to proposals in ACL conferences; instead, we have inclusively considered all datasets employed in RE, irrespective of their origin. The summary of our findings can be referenced in Table \ref{tab:sumresults}.

\begin{table}[H]
    \centering
    \begin{tabular}{lr}
        \hline
        \textbf{Category} & \textbf{Count} \\
        \hline
        Total & 137 \\
        Dataset Papers & 56 \\
        Research Papers (ACL Conferences) & 81 \\
        Passed Inclusion Criteria & 65 \\
        Failed Inclusion Criteria & 16 \\
        \hline
    \end{tabular}
    \caption{Survey summary}
    \label{tab:sumresults}
\end{table}

\section{Background}
\label{sec:background}

This section focuses on the theoretical principles that form the basis of our survey. We aim to provide the reader with the necessary conceptual foundations for RE and Language Models.

\subsection{Relation Extraction}
\label{sec:re}

RE is a task in natural language processing that aims to identify and classify the relationships between entities mentioned in the text. This task can be divided into three main components: named entity recognition, relation identification, and relation classification.

Named Entity Recognition (NER) is a critical initial step in relation extraction. It involves identifying and categorizing named entities within a text, such as the names of individuals, organizations, locations, and other specific entities \cite{sun2018overview}. For example, in the sentence ``Apple Inc. is headquartered in Cupertino, California,'' NER identifies ``Apple Inc.'' as an organization and ``Cupertino, California'' as a location. This step is foundational, determining the entities between which relationships will be analyzed.

Relation identification follows NER and involves detecting pairs of entities related to each other in the text. This task can be accomplished using various techniques, including rule-based approaches or machine learning models \cite{ma-etal-2021-sent}. For instance, in the sentence ``Steve Jobs co-founded Apple Inc.,'' relation identification would recognize the entity pair` `Steve Jobs'' and ``Apple Inc.'' and determine their relationship.

Relation classification is the subsequent process where a predefined label is assigned to each identified entity pair, indicating the nature of their relationship. For example, in the sentence ``Barack Obama was born in Hawaii,'' the entity pair ``Barack Obama'' and ``Hawaii'' is classified with the relationship label ``PlaceOfBirth'' \cite{pouran-ben-veyseh-etal-2020-exploiting}. This task involves categorizing the relationship as one of several predefined types, such as ``PlaceOfBirth,'' ``WorksFor,'' or ``LocatedIn,'' based on the context of the entities involved.

\subsection{Approaches and challenges in Relation Extraction}

RE poses several challenges that are currently being addressed and require further research. Our study we have identified and we will specifically focus on the advancements in RE within the following areas:

\begin{itemize}
    \item \textbf{Document-level Relation Extraction}:  In specific scenarios, relationships between entities extend beyond individual sentences, spanning entire documents. Document-level RE tackles this challenge by considering relationships in a broader context. This task is crucial for applications where understanding connections across multiple sentences or document sections is essential for accurate information extraction.

    \item \textbf{Sentence-level Relation Extraction}: Conversely, sentence-level RE focuses on identifying relationships within individual sentences. This level of granularity is relevant for tasks where relationships are localized and can be adequately captured within the confines of a single sentence. This task is essential for applications with short-form content or when the relations of interest are contained within discrete textual units.

    \item \textbf{Multimodal Relation Extraction}:  With the increasing prevalence of multimodal data, RE extends beyond textual information to include visual cues. Multimodal RE integrates text and visual data to enhance understanding of relationships. This task is relevant in domains where textual and visual information jointly contribute to the overall context, such as image captions, medical reports, or social media content.

    \item \textbf{Multilingual Relation Extraction}: In multilingual relation extraction, the challenge lies in extracting relations and training systems in domains where multiple languages are present. This task includes scenarios where documents or texts may contain information in different languages, requiring models to understand and extract relations effectively across language barriers. This task is particularly relevant in diverse linguistic contexts or global applications where information is presented in multiple languages.

    \item \textbf{Few-Shot and Low-Resource Relation Extraction}: Few-shot learning addresses scenarios with limited labeled data for training RE models. It aims to develop models that generalize well even with minimal labeled examples. In low-resource settings, such as specialized domains like biology or medicine, the scarcity of annotated data poses a significant challenge. The high annotation costs in these domains hinder the availability of large labeled datasets, making it challenging to train accurate RE models using traditional supervised learning methods.

    \item \textbf{Distant Relation Extraction}: Distant RE involves extracting relations between entities based on distant supervision. In this approach, heuristics or existing knowledge bases automatically label large amounts of data. However, this introduces challenges related to noisy labels and the potential inaccuracies in the automatically generated annotations. Addressing these challenges is crucial for improving the reliability of models trained on distantly supervised data.

    \item \textbf{Open Relation Extraction}: OpenRE is dedicated to unveiling previously unknown relation types between entities within open-domain corpora. The primary objective of OpenRE lies in the automated establishment and continual evolution of relation schemes for knowledge bases, accomplished through identifying novel relations within unsupervised data. Within the realm of OpenRE methods, we can find two prominent categories: tagging-based methods and clustering-based methods. 
\end{itemize}

\subsection{Language Models}

LMs process sequences of tokens, denoted as $w = \{w_1, w_2, ..., w_N\}$ , by assigning a probability $p(w)$ to the sequence and incorporating information from previous tokens. This is captured mathematically as follows:

\begin{equation}
p(w) = \prod_t p(w_t \mid w_{t-1}, w_{t-2}, ..., w_1)
\end{equation}

Although $p(w)$ has been traditionally calculated statistically, recently  $p(w)$ through neural language models \cite{Bengio2001}. However, the methods employed to calculate $p(w)$ vary across applications, diverse architectures from multi-layer perceptrons \cite{Mikolov2012} and convolutional neural networks \cite{pmlr-v70-dauphin17a} to recurrent neural networks \cite{Zaremba2014}.

Recently, the introduction of self-attention mechanisms \cite{parikh-etal-2016-decomposable} has heralded the development of high-capacity models, with transformer architectures \cite{NIPS2017_7181} emerging as frontrunners due to their efficient utilization of extensive data, leading to state-of-the-art results. Transformers have revolutionized the field of NLP, demonstrating their potential across various downstream applications, such as question answering \cite{nassiri2023transformer} and text classification \cite{tezgider2022text} or text generation \cite{luo2022biogpt}. Within the transformer architecture, there are different types, with two of the most widespread being the encoder-only and the encoder-decoder models \cite{cai2022compare}.

Encoder-only models are designed to process input data and generate rich, contextualized representations. These models excel at tasks that involve understanding \cite{kadlec-etal-2016-text} and analyzing text, such as named entity recognition. The key feature of encoder-only models is their focus on generating deep contextual embeddings of the input sequence, which capture intricate semantic and syntactic information without producing any output sequence.

On the other hand, Encoder-decoder models are structured to handle tasks that require both understanding and generation of text. This architecture consists of two distinct components: an encoder and a decoder. The encoder processes the input sequence and generates a set of intermediate representations that encapsulate the input's contextual information. The decoder then takes these representations and generates the final output sequence. This design is particularly effective for tasks that involve complex text generation, such as machine translation \cite{cho-etal-2014-properties}, where the model translates text from one language to another, and text summarization \cite{lebanoff-etal-2018-adapting}. The encoder-decoder structure allows for a dynamic interplay between understanding and generation, making it well-suited for applications that require producing coherent and contextually appropriate output based on the input.


Noteworthy is Bi-directional Encoder Representations from Transformers (BERT) \cite{Devlin2018}. BERT employs a unique approach that involves learning the probability $p(w_i)$ by masking parts of the input sequence and predicting the masked word $w_i$ while considering the surrounding context. Specifically, BERT and other contextual transformer-based models aim to compute the probability of a word given its context by estimating:

\begin{equation}
p(w_i) = p(w_i \mid w_1, \ldots, w_{i-1}, w_{i+1}, \ldots, w_N),
\label{eq:2}
\end{equation}

Being $w_i$  the target word to be predicted, and $\{w_1, \ldots, w_{i-1}, w_{i+1}, \ldots, w_N\}$ represents the surrounding context words. This approach allows BERT to capture the contextual dependencies within a given sequence effectively.

The computation of these probabilities is made possible by training the transformer architecture on large and diverse datasets. Specifically, models like BERT are trained on extensive corpora, including the BookCorpus \cite{10.1109/ICCV.2015.11} and a comprehensive crawl of English Wikipedia. This broad training allows the models to learn from vast amounts of text and diverse text combinations, enabling them to capture a wide range of linguistic patterns and contextual nuances. Notably, BERT has demonstrated remarkable prowess in tasks like RE \cite{roy2021incorporating} and knowledge base completion \cite{yao2019kg,lan2021path}, showcasing its ability to capture intricate relationships within the text.

Variations of BERT, such as RoBERTa, have further refined the architecture by leveraging large-scale pretraining and training procedures, enhancing both efficiency and performance. RoBERTa \cite{liu2019roberta} achieves state-of-the-art results through techniques like dynamic masking and larger batch sizes.

One of the most innovative models in this field is called the Generative Pre-trained Transformer (GPT), developed by Radford et al. \cite{radford2018improving} and the Text-to-Text Transfer Transformer (T5) as elucidated by Raffel et al. \cite{raffel2020exploring}.

These models have introduced new ways of understanding language beyond traditional methods. GPT, for instance, distinguishes itself by its remarkable ability to generate coherent and contextually rich text, a feat that has revolutionized text generation tasks. Conversely, T5 adopts a distinct approach by treating many NLP challenges as a unified text-to-text problem. This approach showcases its extraordinary versatility in handling diverse functions within a comprehensive framework.

These language models, each with its unique approach, collectively embody the essence of modern NLP, where theoretical advancements converge with pragmatic utility, opening avenues for extracting nuanced relations and semantic nuances from textual data.

\section{Datasets for RE}
\label{sec:datasets}
We have compiled the most comprehensive datasets used for RE to date. We begin by providing brief descriptions of each dataset. Furthermore, in Section \ref{sec:rq3}, we describe in more detail the most frequently used datasets for RE. We have categorized datasets into two distinct periods. The first comprises cutting-edge datasets introduced during our review period from 2019 to 2023, while the second includes traditional datasets proposed before this time frame.

\subsubsection{Datasets proposed during the survey period}

\begin{itemize}
    \item \textbf{DialogRE}: A dataset derived from dialogues in the Friends TV series, comprising 10,168 subject-relation-object tuples, covering diverse relations such as \textit{`positive\_impression'} and \textit{`siblings'}. The dataset utilizes BERT-base-uncased with special tokens to demarcate subject and object boundaries within dialogues \cite{yu-etal-2020-dialogue}.
    
    \item \textbf{BioRED}: A biomedical relation extraction dataset featuring multiple entity types (e.g., gene, protein, disease, chemical) and relation pairs (e.g., \textit{`gene-disease'},  \textit{`chemical-chemical}) at the document level, tagged on a set of 600 PubMed abstracts, comprising 4,178 relations and 13,351 entities \cite{luo2022biored}.
    
    \item \textbf{Re-DocRED}: An updated version of the DocRED dataset with re-annotation of 4,053 documents to address and rectify issues with false negatives due to incomplete annotation in the original dataset \cite{tan2022revisiting}.
    
    \item \textbf{BioRelEx}: A dataset comprising over 2,000 sentences from biological journals with comprehensive annotations of entities such as proteins, genes, and chemicals, and detailed information about their binding interactions \cite{khachatrian-etal-2019-biorelex}.
    
    \item \textbf{FOBIE}: A dataset focusing on biological relations between elements, including relations such as \textit{`trade-off'}, \textit{`argument-modifier'}, and \textit{`not-a-trade-off'}. It is automatically annotated using a rule-based system and employs word embedding encoding rather than larger language models like BERT \cite{kruiper-etal-2020-laymans}.
    
    \item \textbf{FUNSD}: A dataset for form understanding in noisy scanned documents, comprising 199 fully annotated scanned forms addressing tasks including entity labeling and relation extraction \cite{jaume2019funsd}.
    
    \item \textbf{SIMILER}: A dataset consisting of 1.1 million annotated Wikipedia sentences representing 36 relations and 14 languages, created specifically for entity and relation extraction tasks \cite{seganti-etal-2021-multilingual}.
    
    \item \textbf{HISTRED}: A historical document-level relation extraction dataset constructed from Yeonhaengnok, a collection of historical records with bilingual annotations for various named entities and relations \cite{yang-etal-2023-histred}.
    
    \item \textbf{MultiTACRED}: A multilingual version of the TACRED dataset covering 41 person- and organization-oriented relation types across 12 target languages \cite{hennig-etal-2023-multitacred}.
    
    \item \textbf{TACRED-Revisited / TACREV}: A label-corrected version of the TACRED dataset, addressing challenging cases by carefully validating the most difficult 5,000 examples from the original dataset \cite{alt-etal-2020-tacred}.
    
    \item \textbf{RE-TACRED}: A comprehensive analysis and re-annotation of the entire TACRED dataset \cite{alt2020tacred}.
    
    \item \textbf{REDfm}: A filtered and multilingual relation extraction dataset divided into two subsets: SREDfm, which includes over 40 million triplet instances across 18 languages, and REDfm, a smaller human-reviewed dataset focusing on seven languages \cite{huguet-cabot-etal-2023-red}.
    
    \item \textbf{NMRE}: A multimodal named entity recognition dataset containing text, image, and entity annotations, with 9,201 text-image pairs and 15,485 entity pairs across 23 relation categories \cite{zheng2021multimodal}.
    
    \item \textbf{RISeC}: A dataset focused on procedural text within the cooking domain, offering explicit labels for various relations, including temporal relations and manner descriptions \cite{jiang2020recipe}.
    
    \item \textbf{EFGC}: A dataset specializing in cooking with co-reference relations segmented by tools, foods, or actions \cite{yamakata2020english}.
    
    \item \textbf{WIKIDISTANT}: A dataset derived from Wikipedia, composed of 454 target relations with 1,050,246 elements in training, 29,145 in validation, and 28,897 for testing \cite{han-etal-2020-data}.
    
    \item \textbf{DrugCombination-Dataset}: A dataset designed for relation extraction involving diverse drug combinations, comprising 1,634 biomedical abstracts and expert-annotated to extract information regarding drug combination efficacy \cite{tiktinsky-etal-2022-dataset}.
    
    \item \textbf{DWIE}: The Deutsche Welle corpus for Information Extraction (DWIE) contains 501,095 tokens from news articles, illustrating 317,204 relations across 65 types, with annotations for named entities, co-reference resolution, and relation extraction \cite{zaporojets2021dwie}.
\end{itemize}

\subsubsection{Datasets proposed before the survey period}

\begin{itemize}

\item \textbf{DocRED}: A comprehensive document-level relation extraction dataset, DocRED, is constructed using Wikipedia data. This dataset encompasses two subsets, one manually annotated and the other annotated through distant supervision. DocRED includes 96 diverse relations, including notable ones like \textit{educated\_at} and \textit{spouse}. The dataset comprises 155,535 relation mentions \cite{yao-etal-2019-docred}.
    
    \item \textbf{SciERC}: A dataset designed for constructing scientific knowledge graphs, covering multi-task annotation for entities, relations, and coreference resolution \cite{luan-etal-2018-multi}.
    
    \item \textbf{TACRED}: A large-scale dataset annotated via Amazon Mechanical Turk, focusing on common relations between people, organizations, and locations, emphasizing slot filling \cite{zhang-etal-2017-position}.
    
    \item \textbf{NYT-10/FB-NYT}: A dataset addressing relation modeling without explicitly labeled text, exploring relationships mentioned in news articles \cite{riedel2010modeling}.
    
    \item \textbf{BioCreative V CDR Task Corpus}: A chemical-disease relation extraction resource contributing to biomedical research \cite{li2016biocreative}.
    
    \item \textbf{ChemProt}: A dataset that extracts various chemical-protein interactions within the biomedical domain \cite{kim2012chemprot}.
    
    \item \textbf{GDA}: The corpus on gene-disease associations comprises 30,192 titles and abstracts extracted from PubMed articles. These entries have been annotated to identify genes, diseases, and gene-disease associations through distant supervision. A subset of 1000 examples from this corpus forms the test set. The relationships within the corpus are rooted in the genetic association database \cite{bravo2015extraction,becker2004genetic}.
    
    \item \textbf{FewRel}: A large-scale dataset for few-shot relation classification featuring 100 relations with training, validation, and testing splits \cite{han-etal-2018-fewrel}.
    
    \item \textbf{Wiki-ZSL}: A zero-shot learning benchmark for relation extraction utilizing Wikipedia articles, including seen and unseen relations \cite{levy-etal-2017-zero}.
    
    \item \textbf{Wiki80}: A dataset with 80 relation types designed for evaluating neural relation extraction models \cite{han2019opennre}.
    
    \item \textbf{ACE05}: A multilingual training corpus used in the 2005 Automatic Content Extraction (ACE) technology evaluation \cite{walker2006ace}.
    
    \item \textbf{SemEval-2010 Task8}: A benchmark dataset for relation extraction, focusing on multi-way classification of semantic relations between pairs of nominals \cite{hendrickx2019semeval}.
    
    \item \textbf{SKE}: A dataset tailored for relation extraction from the Chinese social network Baidu \footnote{http://ai.baidu.com/broad/download?dataset=sked}.
    
    \item \textbf{ERE}: A dataset facilitating comparison of events and relations across different annotation standards \cite{aguilar2014comparison}.
    
    \item \textbf{WebNGL}: Crafted using micro-planning data-to-text corpora extracted from existing Knowledge Bases, namely DBpedia, encompassing a total of 21,855 pairs of data and text \cite{gardent2017creating}.
    
    \item \textbf{ADE}: A benchmark corpus for the automated extraction of drug-related adverse effects from medical case reports. The documents undergo a double-annotation process across multiple rounds to ensure uniform and reliable annotations \cite{gurulingappa2012development}.
    
    \item \textbf{GIDS}: A dataset with 50,000 examples addressing relations related to institutions, places of birth, and attended locations \cite{jat2018improving}.
    
    \item \textbf{Twitter 15}: A dataset containing text and image pairs from Twitter posts utilized for multimodal named entity recognition and relation extraction tasks \cite{lu2018visual}.
    
    \item \textbf{Twitter 17}: A dataset comprising text and image pairs from Twitter posts, used for multimodal named entity recognition and relation extraction tasks \cite{zhang2018adaptive}.
    
    \item \textbf{MSCorpus}: A dataset concentrating on materials science synthesis procedures, defining a labeling schema specialized for this domain, including various relations \cite{mysore2019materials}.
    
    \item \textbf{CoNLL}: A bilingual (German and English) dataset specifically designed for named entity recognition \cite{tjong-kim-sang-de-meulder-2003-introduction}.
    
    \item \textbf{EventStoryLine}: A dataset designed for narrative understanding and event extraction \cite{mostafazadeh2016caters}.
    
    \item \textbf{Causal-TimeBank}: A dataset for relation extraction based on time series focusing on causal relations within the TimeBank corpus \cite{mirza2014analysis}.
    
    \item \textbf{Matres}: A dataset for temporal relation extraction based on time series \cite{ning2018multi}.
    
    \item \textbf{RE-QA}: A dataset with ten folds containing 84 relation types in the train, 12 on the dev, and 24 on the test splits. There is no overlap among the relation types of these splits per fold. Each fold contains 840k sentences in the train, 6k in the dev, and 12k in the test split. Half the sentences are negative examples, and gold questions are provided for each relation type \cite{levy-etal-2017-zero}.
    
    \item \textbf{KBP37}: A sentence-level relation extraction dataset with 21,046 examples collected from 2010 and 2013 KBP documents and the July 2013 dump of Wikipedia \cite{zhang2015relation}.
\end{itemize}


\section{Cutting-edge RE techniques based on language models}
\label{sec:review}

We employed a two-fold approach to facilitate a thorough analysis of the field's evolution. First, we conducted a task-based analysis, offering a comprehensive examination of the techniques used to address each specific task in relation extraction. This structured approach provides a detailed understanding of how various methods have been applied and evolved across different RE tasks. Second, we organized our survey into chronological tables by year, conference, task, sub-task, and model type. This segmentation enables us to explore the progression of techniques over time, identify emerging trends, and derive valuable insights from the evolving landscape of RE methods.

\subsection{Task analysis of cutting-edge RE}

In this section, we will discuss how new systems and models are addressing the tasks and challenges of relation extraction (RE) that were introduced in Section \ref{sec:re}. We will particularly emphasize the systems and models that make use of language models (LMs) and large language models (LLMs).

\subsubsection{Sentence-level Relation Extraction} \hfill\\

In\cite{veyseh2020exploiting}, Veyseh et al. used a Ordered-Neuron Long-Short Term Memory Networks (ON-LSTM) to infer model-based importance scores for every word within sentences. This paper uses dependency trees to calculate importance scores for sets of words based on syntax. These syntax-based scores are then regulated to align with the model-based importance scores, ushering in a harmonious synergy between syntax and semantics. This approach achieved state-of-the-art performance levels on three distinguished RE benchmark datasets: ACE 2005, related to news; SPOUSE with only one entity type and relation, related to whether two entities regarding people are married; and SciERC, related to scientific abstracts. They used BERT to obtain pre-trained word embeddings for the sentences. Specifically, they use the hidden vectors in the last layer of the BERT-base model (with 768 dimensions) to obtain the pre-trained word embeddings for the sentences. They fixed BERT in the experiments and compared the performance of their proposed model with BERT embeddings against other models that used word2vec embeddings. In this paper, word embeddings achieve state-of-the-art results, so it is not surprising that there will be a stream in 2020 that harnesses the potential of word embeddings. 

In \cite{yu-etal-2020-dialogue}, Yu et al. introduced DialogRE, a dataset derived from dialogues in the Friends TV series. The dataset contains 10,168 subject-relation-object tuples, encompassing a wide range of relations such as ``positive\_impression'' and ``siblings''. The paper harnesses BERT-base-uncased, employing unique tokens to demarcate subject and object boundaries within the dialogues. Notably, this work pioneers RE in dialogue contexts, conducting experiments against traditional CNN, LSTM, and BiLSTM-based models, often coupled with the GloVe encoder.

In \cite{yuan-eldardiry-2021-unsupervised}, the authors explore unsupervised relation extraction, a technique that does not require labeled data to identify relations between entities in text. The paper proposes a new approach that overcomes the limitations of existing variational autoencoder-based methods. The proposed method improves training stability and outperforms state-of-the-art methods on the NYT dataset. The authors use a variational autoencoder to learn a low-dimensional representation of sentences that captures their relation information. They also introduce a novel regularization term, encouraging the model to learn more informative representations. The proposed approach is evaluated on the NYT dataset and compared to four state-of-the-art methods. The results show that the proposed method achieves higher F1 scores than the baselines. HiURE \cite{liu-etal-2022-hiure} advances unsupervised relation extraction by introducing a contrastive learning framework. This approach leverages hierarchical exemplars to enhance the relational features of sentences. It optimizes exemplar-wise contrastive learning through iteratively integrating HiNCE and propagation clustering. This method stands out from existing unsupervised RE models, as it avoids gradual drift and surpasses instance-wise contrastive learning by actively considering the hierarchical structure inherent in relations.

One of the most debated topics in RE is the choice between pipeline frameworks and joint methods. As discussed in Section \ref{sec:re}, RE involves three sub-techniques. Pipeline frameworks tackle individual tasks in RE, such as entity recognition, relation identification, and relation classification, separately, while joint methods integrate these tasks into a unified model. This distinction raises questions about which approach is more effective for improving performance and efficiency in relation extraction systems. While most approaches treat each task as a separate problem, Wang et al. \cite{wang-etal-2021-unire} propose a joint system for entity detection and relation classification. The key idea is to view entity detection as a particular case of relation classification. The authors introduce a new input space, a two-dimensional table with each entry corresponding to a word pair in sentences. The joint model assigns labels to each cell from a unified label space. After filling the table, the authors apply a Biaffine layer to predict the final relation labels. The proposed UNIRE model is based on BERT-base-uncased, ALBERT, and SciBERT as encoding depending on the dataset and achieves state-of-the-art performance on the ACE04 and ACE05 datasets, outperforming several baseline models.

\cite{zhong-chen-2021-frustratingly},introduce a pipeline framework for RE that capitalizes on the results obtained from language models to create a straightforward and highly accurate approach. This approach entails employing distinct models for entity recognition and relation extraction, with the latter model taking entity pairs as its input. The models are trained independently, utilizing cross-sentence context to enhance overall performance. For each task, the authors harnessed the power of pre-trained language models, specifically BERT, ALBERT, and SCIBERT, constructing two separate encoders for entity recognition and relation extraction. Both tasks saw significant performance improvements by training entity and relation models independently and having the relation model use the entity model for input features.

The authors in \cite{yan-etal-2022-empirical} compared the performance of pipeline and joint approaches to Entity and Relation Extraction (ERE) tasks. The research focused on two specific datasets, ACE2005 and SciERC, which are widely used for evaluating information extraction methods. The authors used eight different methods, including purely sequential methods and four combined models, to see how well they could extract entities and relationships from the text. Their experiments used two pre-trained language models: BERT-base-uncased for the ACE2005 dataset and Seibert-sci-vocab-uncased for the SciERC dataset. The study revealed that while pipeline approaches can yield competitive results, the best-performing joint approach consistently outperformed the best pipeline model across various metrics. This finding underscores the potential advantages of joint models, particularly in leveraging the interdependencies between tasks. The ERE tasks performed in this study operate primarily at the sentence level, extracting entities and relations from individual sentences within the documents.

 Jie et al. \cite{jie-etal-2022-learning} introduced an approach for solving math word problems, which can be considered a sentence-level task. Their method employs explainable deductive reasoning steps to construct target expressions iteratively. Each step involves a primitive operation over two quantities, defining their relation. By framing the task as a complex RE problem, their model significantly outperforms existing baselines in accuracy and explainability. The primary objective is to identify relations between numbers, where mathematical expressions represent these relations. The study uses three English and one Chinese dataset, outperforming state-of-the-art methods established by previous works \cite{koncel2016parsing,wang2017deep,amini2019mathqa,patel2021svamp}. The authors leverage BERT as the encoder for their proposed model. Specifically, they utilize Chinese BERT and Chinese RoBERTa for the Math23k dataset and BERT and RoBERTa for the English datasets. The experimentation involves also multilingual BERT and XLM-RoBERTa. The pre-trained models are initialized from HuggingFace's Transformers, with the best-performing model identified as related to RoBERTa.

In Wang et al. \cite{wang-etal-2022-rely}, the authors delve into the issue of entity bias in sentence-level relation extraction. The paper's main contribution is the introduction of CORE (Counterfactual Analysis Relation Extraction): a model-agnostic technique explicitly designed to address this problem. The authors showcase that CORE effectively enhances the efficacy and generalization of prevalent RE models by distilling and mitigating biases entity-awarely. The approach involves formulating existing RE models as causal graphs and applying counterfactual analysis to post-adjust biased predictions during inference. Language models, particularly RoBERTa, are significant because of the flexibility with which CORE can be seamlessly integrated into popular RE models to boost their performance and mitigate biases without necessitating retraining. In contrast to approaches incorporating graphs or additional information, \cite{liang-etal-2022-modeling} introduced an innovative method for RE that exclusively extracts multi-granularity hierarchical features from the original input sentences. The model employs hierarchical mention-aware segment attention and global semantic attention to capture features at various levels. These features are then aggregated for relation prediction. While emphasizing the effectiveness of BERT and SpanBERT \cite{joshi2020spanbert} as encoders, the authors also incorporate a BiLSTM model.

The authors in\cite{zhou-chen-2022-improved} leveraged RoBERTa as the foundational framework for enhancing the baseline model in sentence-level relation extraction. The paper presents a dual-fold contribution, marked by two advancements. Firstly, the authors introduce an improved baseline specifically tailored for sentence-level relation extraction, strategically tackling two prominent challenges prevalent in existing models: effective entity representation and handling noisy or ill-defined labels. Secondly, the study establishes and underscores the efficacy of pre-trained language models, particularly RoBERTa, in sentence-level relation extraction. This efficacy is substantiated by achieving state-of-the-art results on the TACRED dataset and its refined iterations. Notably, the proposed model stands out as a solution that adeptly addresses two critical issues impacting the performance of contemporary RE models: entity representation and dealing with noisy or ill-defined labels.

In 2022, large language models such as GPT became mainstream. Yang and Song \cite{yang-song-2022-fpc} introduced a prompt-based fine-tuning approach to enhance relation extraction. The foundation of their method involved utilizing RoBERTa large as the base model, subjected to fine-tuning on the training sets of respective datasets through the prompt-based fine-tuning methodology. Throughout the fine-tuning process, the authors employed prompts featuring a masked relation label, tasking the model with predicting the obscured label based on the input sentence. A carefully crafted prompt learning curriculum was also implemented to facilitate the model's adaptation to a multi-task setting, introducing tasks of increasing difficulty. The experimental findings underscore the efficacy of the proposed model. Notably, the model attained state-of-the-art results on benchmark datasets, namely TACRED and SemEval 2010. These outcomes affirm the method's prowess in advancing the state-of-the-art in RE tasks.

\subsubsection{Document-level Relation Extraction}\hfill\\

Document-level RE's most discussed topics are evidence selection, joint and graph models, and new training strategies.

Evidence selection refers to identifying and selecting relevant sentences or text from a document that provides crucial information to determine relationships between entities. In \cite{huang-etal-2021-entity}, the authors present E2GRE, a model designed for document-level relation extraction. E2GRE tackles the challenge of handling relations spanning multiple sentences by jointly extracting relations and their supporting evidence sentences using BERT as an input encoder. BERT plays a central role in the E2GRE framework by contextualizing document text and entity mentions. The last four layers of BERT are used for RE and evidence prediction. Notably, E2GRE enhances BERT's attention mechanism to focus on relevant context, using attention probabilities as supplementary features for evidence prediction. While E2GRE achieves state-of-the-art results in evidence prediction on the DocRED dataset, it needs to attain the same level of accuracy in relation classification. The model uses RoBERTa \cite{liu2019roberta} as an input encoder, broadening its capabilities in document-level relation extraction.

In \cite{nan-etal-2020-reasoning}, the authors introduce an approach to document-level relation extraction. This model sets itself apart from existing methods in the field by employing a refinement strategy that progressively aggregates pertinent information to facilitate multi-hop reasoning. This distinctive strategy, Latent Structure Refinement (LSR), contrasts traditional models reliant on fixed syntactic trees. Instead, LSR dynamically learns the document's structure and utilizes this acquired knowledge to inform predictions. The model's underlying encoding mechanisms draw from both GloVe and BERT, allowing it to achieve state-of-the-art results when applied to datasets such as CDR  \cite{li2016biocreative}, GDA \cite{han2016global}, and DocRED \cite{yao-etal-2019-docred}, especially when employing BERT encoding.  

Similarly to the preceding paper, Huang et al. \cite{huang-etal-2021-three} propose a system for document-level RE that focuses on selecting evidence sentences rather than performing RE directly. The authors present a method for heuristically selecting informative sentences from a document that can be used as evidence for identifying the relationship between a given entity pair. The selected evidence sentences can then be input into a BiLSTM-based RE model. The authors show that their method for evidence sentence selection outperforms graph neural network-based methods for document-level RE on benchmark datasets, including DocRED, GDA, and CDR.

Xu \cite{xu-etal-2022-document} introduces a paper that complements traditional document-level RE tasks . The framework, named SIEF, can be integrated with established approaches like BERT. SIEF enhances the performance of DocRE by introducing importance scores to sentences, directing the model's attention to evidence-rich sentences. This approach fosters consistent and robust predictions. The methodology involves calculating sentence importance scores and implementing a sentence-focusing loss to promote model robustness. The study underscores the synergistic relationship between SIEF and modern language models, such as BERT.

Document-level relation extraction has also been explored from a joint modeling perspective. Eberts and Ulges \cite{eberts-ulges-2021-end} present a comprehensive approach to entity-level RE that covers all stages of the RE process. Their proposed joint model extracts mentions, clusters them into entities, and classifies relations jointly. The joint model can leverage shared parameters and training steps across the different sub-tasks, which improves efficiency and performance compared to a pipeline approach where each sub-task is trained separately. The authors also use a multi-instance learning approach that aggregates information from multiple mentions of the same entity to improve performance. The model incorporates BERT to encode contextual information. The proposed joint model achieves state-of-the-art results on the challenging DocRED dataset, demonstrating the power of joint modeling for entity-level RE. 

\cite{xu-choi-2022-modeling} introduced COREF, a Graph Compatibility (GC) approach rooted in the bidirectional interaction between coreference resolution and RE. The GC method capitalizes on task-specific traits to actively influence the decisions of distinct tasks, establishing explicit task interactions that avoid the isolated decoding of each task. The authors employed SpanBERT to capture intricate contextual dependencies for encoding, 

Zhang et al. \cite{zhang-etal-2023-novel} propose a system, TaG, adopts a joint approach grounded in graph-based methodologies, incorporating table filling, graph construction, and information propagation techniques. To capture rich contextual information, the authors employ BERT and RoBERTa as encoders over the tables, with particular success achieved through leveraging RoBERTalarge.


Document-level relation extraction involves identifying and classifying relationships between entities within entire documents, which presents unique challenges compared to sentence-level extraction. Traditional training and evaluation techniques often fall short because they may need to adequately capture the complexities of long-range dependencies and contextual nuances inherent in document-level data. Standard methods might focus on sentence-by-sentence analysis, overlooking the importance of the document's overall structure and coherence. Consequently, new training techniques are required to effectively manage the large-scale and multi-sentence contexts in which entities are situated.

Xiao et al., in \cite{xiao-etal-2022-sais},  introduced SAIS, an approach employing language models, such as BERT, RoBERTa, and SCIBERT, to encode extensive contextual dependencies. This strategic use of language models addresses the complexities of capturing extended contexts and nuanced interactions among entities in document-level relation extraction. The primary contribution of the SAIS method lies in its explicit guidance for models to adeptly capture pertinent contexts and entity types during document-level relation extraction. This emphasis results in enhanced extraction quality and more interpretable predictions from the model.

Chen et al. \cite{chen-etal-2023-models} introduce a novel perspective by evaluating models based on language understanding and reasoning capabilities. Employing the Mean Average Precision (MAP) metric and subjecting models to RE-specific attacks, the authors shed light on significant disparities in decision rules between state-of-the-art models and human approaches. Notably, these models often depend on spurious patterns while overlooking crucial evidence words, adversely affecting their robustness and generalization in real-world scenarios. Although the paper does not propose new models for RE and does not explicitly use language models, it aligns with our research goal of understanding how language models capture information for relation extraction. Instead, the authors compare and evaluate various baselines, including BERT and RoBERTa, providing valuable insights into their comparative effectiveness. 

In 2023, Guo et al. \cite{guo-etal-2023-towards} introduced PEMSCL, to enhance relation prediction accuracy by integrating discriminability and robustness. The approach employs a pairwise moving-threshold loss with entropy minimization, incorporates adapted supervised contrastive learning, and introduces a novel negative sampling strategy. For encoding, the authors used ATLOP \cite{zhou2021document}. Ma et al. \cite{ma-etal-2023-dreeam} introduced a method called DREEAM. This approach uses evidence information to guide the attention modules to emphasize important evidence. One notable aspect is that the model is self-trained to learn relationships between entities using automatically generated evidence from large amounts of data without requiring explicit evidence annotations. This approach's encoder for evidence retrieval is RoBERTa, which significantly contributes to the system's overall effectiveness. DREEAM currently stands as the most accurate system for DocumentRE.

In \cite{zhao-etal-2023-improving}, a continual document-level relation extraction model is introduced to address the challenges of distinguishing analogous relations and preventing overfitting to stored samples. The model employs a learning framework that integrates a contrastive learning objective with a linear classifier training one. It generates memory-insensitive relation prototypes by combining static and dynamic representations, which helps maintain robustness against noise from stored samples. The model uses memory augmentation to create more training samples for replay, enhancing its adaptability in continual learning scenarios. Focal knowledge distillation is also employed to assign higher weights to analogous relations, ensuring the model focuses on subtle distinctions between similar relations. For encoding, the authors use BERT, leveraging its robust contextual embeddings.


Zhang et al. (2021) discussed the challenges of extracting information from visually rich documents (VRDs) in their paper \cite{zhang-etal-2021-entity}. The complexity arises from the need to incorporate both visual and textual features in the extraction process. The authors propose an approach that adapts the affine model used in dependency parsing to the entity RE task and conduct experiments on the FUNSD dataset and a real-world customs dataset to compare different representations of the semantic entity, different VRD encoders, and different relation decoders. They also employ two training strategies, multi-task learning with entity labeling and data augmentation, to further improve their model's performance. The proposed model outperforms previous baselines on the FUNSD dataset, and achieves high performance on the real-world customs dataset. The authors also analyze the performance of different language models, such as BERT and LayoutLM, on their language mixed data and find that encoding layout information into language models significantly enhances the model's ability to understand the spatial relationships and contextual relevance of entities, leading to improved accuracy in relation extraction tasks.

Finally, Yuan et al. \cite{yuan-etal-2023-discriminative} present SENDIR, a model specifically tailored to tackle the complexities of document-level reasoning in event-event relation extraction. This model introduces an approach by learning sparse event representations, enabling effective discrimination between intra- and inter-sentential reasoning. The paper addresses the challenges associated with comprehending lengthy texts. Experimental validation of the model was conducted across three datasets related to event detection: EventStoryLine, Causal-TimeBank, and MATRES. The SENDIR model incorporates language models to enhance its capabilities, employing a BERT-base-uncased in conjunction with a BiLSTM.

\subsubsection{Multilingual and multimodal Relation Extraction}\hfill\\

Multilingual relation extraction has benefited from several developments in recent years. In \cite{seganti-etal-2021-multilingual}, the authors introduce a new dataset, SMILER, comprising 1.1 million annotated sentences spanning 14 languages. To tackle this dataset, they propose the HERBERTa model, a BERT-based pipeline integrating two independent BERT models for sequence classification and entity tagging. The authors employ the multilingual BERT model, M-BERT, pre-trained on monolingual corpora in 104 languages and fine-tuned BERT models tailored for specific languages. Notably, the HERBERTa model achieves performance close to state-of-the-art in multilingual relation extraction.

Yang et al. in \cite{yang-etal-2023-histred} introduce a new dataset for historical RE based on the Yeonhaengnok, a collection of historical records written initially in Hanja and later translated into Korean. The dataset contains 5,816 data instances and includes ten types of named entities and 20 relations. The authors propose a bilingual RE model to extract relations from the dataset that leverages Korean and Hanja contexts. The model leverages pre-trained language models, including KLUE, a BERT-based model designed explicitly for Korean natural language processing tasks, and AnchiBERT, tailored for Hanja. By leveraging both Korean and Hanja contexts, the proposed model outperforms other monolingual models on the HistRED dataset, demonstrating the effectiveness of employing multiple language contexts in RE tasks.  \cite{hennig-etal-2023-multitacred} introduces MultiTACRED. This extension of the TACRED dataset spans 12 different languages and aims to explore the intricacies of multilingual relation extraction further. Unlike proposing novel models, the paper focuses on meticulously evaluating monolingual, cross-lingual, and multilingual models. This assessment delves into the performance across all 12 languages the dataset covers. Additionally, the study scrutinizes the effectiveness of pre-trained mono- and multilingual encoders, particularly highlighting the role of BERT-based in tackling challenges posed by multilingual natural language processing tasks and cross-lingual transfer learning scenarios. \cite{huguet-cabot-etal-2023-red} also contributes to multilingual RE by introducing REDfm, a dataset designed for multilingual models. This dataset encompasses various relation types across multiple languages, offering higher annotation coverage and more evenly distributed classes than current datasets. Additionally, the authors propose a multilingual RE model named mREBEL. As an extension of the REBEL model, mREBEL adopts a seq2seq architecture to extract triplets, encompassing entity types across various languages. The authors conducted pre-training for mREBEL using mBART-50 and fine-tuned it on the RED FM dataset. The results demonstrate that mREBEL surpasses the performance of existing models on the REDfm dataset.

In multimodal RE, recent innovations have focused on integrating visual and textual information. In \cite{zheng-etal-2023-rethinking}, authors addressed multimodal entity and RE through an approach that integrates visual and textual information. The model uses techniques such as back-translation and multimodal divergence estimation to exploit a unified transformer. By using convolutional neural networks and BERT-base-uncased as encoders for visual and textual content, the study uses Twitter-2015, Twitter-2017, and MNRE datasets designed explicitly for multimodal tasks, achieving state-of-the-art results. In the same line, \cite{wu-etal-2023-information} proposed a graph-based framework for multimodal RE. The framework involves representing the input image and text with visual and textual scene graphs, which are then fused into a unified cross-modal graph. The graph is then refined using the graph information bottleneck principle to filter out less informative features. The model uses latent multimodal topic features to enhance the feature contexts, and then the decoder uses these enriched features to predict the relation label. The model is pretrained using CLIP (vit-base-patch32), which performed better than BERT on the MRE \cite{zheng2021multimodal} dataset, comprising 9,201 text-image pairs. Hu et al. present in \cite{hu-etal-2023-multimodal} an innovative method that introduces an strategy for synthesizing object-level, image-level, and sentence-level information. This approach enhances the capacity for reasoning across various modalities, facilitating improved comprehension of similar and disparate modalities. The proposed method underwent testing on the MNRE dataset, achieving state-of-the-art results. BERT-base-uncased served as the encoder for textual information.

\subsubsection{Few-Shot and Low-Resource Relation Extraction}\hfill\\

The need to enhance model performance in scenarios where labeled data is scarce has led to the development of specialized techniques such as few-shot, low-resource, and zero-shot learning. These approaches are essential for making RE systems more adaptable and practical, especially in real-world applications where obtaining large, annotated datasets is often challenging or impractical. Few-shot and low-resource RE methods address these limitations by enabling models to learn effectively from minimal data, reducing the dependency on extensive manual annotation. Additionally, the incorporation of external knowledge, innovative training strategies, and advanced semantic matching techniques has enhanced these methods, resulting in more precise and broadly applicable models.

In \cite{yang2021entity}, the authors address the challenge of inaccurate relation classification in low-resource data domains driven by limited samples and a knowledge deficit. They introduced the ConceptFERE scheme, an advancement in few-shot RE algorithms. This scheme incorporates a concept-sentence attention module that selects the most relevant concept for each entity by calculating semantic similarity between sentences and ideas. A self-attention-based fusion module also bridges the gap between concept and sentence embeddings from different semantic spaces. The paper adopts BERT as the foundational model for relation classification. This BERT model undergoes pre-training on a large text corpus and fine-tuning on the FewRel \cite{han-etal-2018-fewrel} dataset. The authors initialize the BERT parameters with BERT-base-uncased.

Brody et al. in \cite{brody-etal-2021-towards} tackle the challenge of performance variability across relation types in few-shot relation classification models. While these models have demonstrated impressive results, their reliance on entity-type information makes distinguishing between relations involving similar entity types challenging. To address this limitation, the authors propose modifying the training routine to enhance the models' ability to differentiate such ties. The suggested enhancements augment the training data with relations involving similar entity types. Through evaluations on the FewRel 2.0 dataset, the paper demonstrates that these modifications substantially improve performance on unseen relations, achieving up to a 24\% enhancement under a P@50 problem (precision with 50 examples). The study utilizes BERT as a pre-trained language model to initialize the few-shot relation classification models.

In \cite{zhao-etal-2023-matching} the authors present a fine-grained semantic matching method tailored for zero-shot relation extraction, explicitly capturing the matching patterns inherent in relational data. Additionally, the paper introduces a context distillation method designed to mitigate the negative impact of irrelevant components on context matching. The effectiveness of the proposed method is evaluated on two datasets: Wiki-ZSL and FewRel. For the encoder model, Bert-base-uncased is employed as the input instance encoder. Comparative assessments are conducted against several state-of-the-art methods, including REPrompt, a competitive seq2seq-based ZeroRE approach that utilizes GPT-2 to generate pseudo data for new relations during fine-tuning. The results demonstrate that the proposed method surpasses the performance of all others in the comparison. In \cite{gururaja-etal-2023-linguistic}, the authors tackle challenges in few-shot RE across domains. The study delves into the influence of linguistic representations, specifically syntactic and semantic graphs derived from abstract meaning representations and dependency parses, on the performance of RE models in few-shot transfer scenarios. Employing BERT-base-uncased, the authors extract embeddings for each entity's constituent tokens and max-pool them into embeddings. These embeddings initialize the feature vectors of nodes in the linguistic graph. The graph-aware model integrates these BERT-derived graph-based features, enhancing RE performance in few-shot scenarios. Validation involves two datasets: one focusing on cooking relations, especially between food and elaborations in recipes, and the other related to materials.

\cite{arcan-etal-2022-towards}, while not directly centered on relation extraction, harnesses the capabilities of RE for constructing a chatbot using a pertinent corpus of language data. The application was tailored to industrial heritage in the 18th and 19th centuries, focusing on the industrial history of canals and mills in Ireland. The authors curated a corpus from relevant newspaper reports and Wikipedia articles, employed the Saffron toolkit to extract pertinent terms and relations within the corpus, and utilized the extracted knowledge to query the British Library Digital Collection and the Project Gutenberg library. Although the paper does not explicitly mention BERT, it marks the initial appearance of the T5 model. The authors suggest that leveraging the capabilities of T5 in future enhancements could lead to further improvements in the system's performance. By utilizing T5's architecture, designed for a wide range of NLP tasks, the model can better capture complex relationships and improve its ability to generalize across different relation types, ultimately addressing the limitations observed in current few-shot models.

In \cite{najafi-fyshe-2023-weakly}, Najafi and Fyshe explore the domain of zero-shot relation extraction. The authors propose an approach that circumvents the need for manually crafted gold question templates by generating questions for unseen relations. The critical advancement lies in the introduction of the OffMML-G training objective. This objective fine-tunes question-and-answer generators specifically tailored for ZeroShot-RE. The result is the generation of semantically relevant questions for the answer module based on the given head entity and relationship keywords. The T5 model serves as the answer generator, having undergone pretraining and fine-tuning on five distinct question-answering datasets. \cite{hu-etal-2021-gradient} presents an approach named Gradient Imitation Reinforcement Learning (GIRL) tailored for Low Resource Relation Extraction. GIRL's primary challenge is extracting semantic relations between entities from text when confronted with a scarcity of labeled data. To address this limitation, GIRL uses gradient imitation to create pseudo labels with fewer biases and errors. It also improves the relation labeling generator within a reinforcement learning framework. The results demonstrate GIRL's superiority over several state-of-the-art methods on benchmark datasets, SemEval and TACRED, achieving competitive performance with significantly less labeled data. The authors employed the BERT default tokenizer with a max-length of 128 for data preprocessing to encode contextualized entity-level representations for the relation labeling generator.

Chen and Li in \cite{chen-li-2021-zs} address zero-shot RE by leveraging text descriptions of both seen and unseen relations to learn attribute vectors as semantic representations. Their strategy enables accurate predictions of unseen ties. Their model outperforms existing approaches in zero-shot settings. The significant influence of BERT on ZS-BERT underscores the power of leveraging sentence-BERT's \cite{reimers2019sentence} contextual representation learning capabilities for encoding input sentences and relation descriptions. Notably, ZS-BERT ranks as the fifth most accurate system for zero-shot relation classification to date, underscoring the enduring impact of BERT as a powerful tool in relation extraction.

The creation of relation prototypes is explored in \cite{han-etal-2021-exploring}. The authors propose few-shot RE (FSRE), an approach based on hybrid prototypical networks and relation-prototype contrastive learning. This method leverages entity and relation information to improve the model's generalization ability. FSRE achieves state-of-the-art results on two datasets, FewRel 1.0 and 2.0, and outperforms existing models by a significant margin. The authors also use BERT as the encoder to obtain contextualized embeddings of query and support instances.

In \cite{liu-etal-2022-pre}, Liu et al. proposed an approach to low-shot RE that unifies zero-shot and few-shot RE tasks. The method, Multi-Choice Matching Networks (MCMN) with triplet-paraphrase pretraining, achieves state-of-the-art performance on three low-shot RE tasks. The datasets used in the experiments are FewRel and TACRED, widely used benchmarks for low-shot RE. The paper assimilates BERT by using it as a backbone model for MCMN and pre-training it with triplet-paraphrase. The experimental results show that MCMN with triplet-paraphrase pretraining outperforms previous methods in all three low-shot RE tasks and achieves state-of-the-art performance.

\cite{qin-joty-2022-continual} investigate a concept called continuous few-shot Relation Learning (CFRL). CFRL is relevant to real-life situations where there is usually enough data available for an established task, ut only a small amount of labeled data for new tasks that come up.. Acquiring large labeled datasets for every new relation is expensive and sometimes impractical for quick deployment. CFRL aims to quickly adapt to new environments or tasks by exploiting previously acquired knowledge. The proposed model is based on embedding space regularization and data augmentation. The authors assimilate BERT and other language models by using them as feature extractors for the input sentences. They show that their method generalizes to new few-shot tasks and avoids catastrophic forgetting of previous tasks. The experiments are conducted on three datasets, and the results show that the proposed method outperforms state-of-the-art methods on all three datasets. The authors also conduct ablation studies to show the effectiveness of each component of their method.

Teru \cite{teru-2023-semi} addresses the challenge of obtaining high-quality labeled data for RE by exploring the influence of data augmentation. The author leverages pre-trained translation models to generate diverse data augmentations for a given sentence in German and Russian. Then \cite{teru-2023-semi} uses lexically constrained decoding strategies to obtain paraphrased sentences while retaining the head and the tail entities. The proposed REMIx system uses a BERT-base-uncased as the encoder. The paper's main contribution is demonstrating the effectiveness of data augmentation and consistency training for semi-supervised relation extraction, achieving competitive results on four benchmark datasets.


Building on previous advances, the creation of realistic benchmarks and domain-specific datasets, such as those focusing on biomedical and document-level RE, has been crucial in testing the applicability of these approaches in diverse, low-resource environments.

Popovic and Farber \cite{popovic-farber-2022-shot} introduced a novel benchmark named FREDo, which is explicitly designed for Few-Shot Document-Level RE (FSDLRE). Unlike existing benchmarks that primarily target sentence-level tasks; the authors claim FREDo provides a more realistic testing ground. The authors proposed two approaches to address FSDLRE tasks that incorporates a modification to the state-of-the-art few-shot RE approach, MNAV. While results demonstrate the superiority of the proposed methods over the baseline built on BERT, they also reveal the ongoing challenges associated with realistic cross-domain tasks, such as domain adaption or the high proportion of negative samples. The study underscores the significance of realistic benchmarks and emphasizes the necessity for substantial advancements in few-shot RE approaches to make them applicable in real-world, low-resource scenarios.

Contributing to the domain of low-resource domains, Tiktinsky et al. \cite{tiktinsky-etal-2022-dataset} introduce an expert-annotated dataset, DrugCombination-Dataset, tailored for the intricate task of N-ary RE involving drug combinations from the scientific literature. To offer more robust results, they conducted a comparative study, pitting various baselines against prominent scientific language models, namely SciBERT \cite{beltagy2019scibert}, BlueBERT \cite{peng-etal-2019-transfer}, PubmedBERT \cite{gu2021domain}, and BioBERT \cite{lee2020biobert}. The results underscore the effectiveness of domain-adaptive pretraining, with PubmedBERT demonstrating the most robust performance among the considered models.

In \cite{xu-etal-2023-nli}, Xu et al. introduced a novel approach called NBR (Natural Language Inference-based Biomedical Relation extraction). NBR addresses the challenges of annotation scarcity and instances without pre-defined labels by converting biomedical RE into a natural language inference formulation, providing indirect supervision, and improving generalization in low-resource settings. The approach also incorporates a ranking-based loss to calibrate abstinent instances and selectively predict uncertain cases. Experimental results demonstrate that NBR outperforms existing state-of-the-art methods on three widely used biomedical RE benchmarks: ChemProt, DDI, and GAD. The paper leverages BioLinkBERTlarge, a BERT variation designed explicitly for biomedical text, to enhance the performance of the proposed NBR approach. By fine-tuning BioLinkBERTlarge on the NBR task, the paper demonstrates the effectiveness of leveraging domain-specific pretraining for improved biomedical RE in low-resource environments. \cite{zhou-etal-2023-continual} proposes a method that pre-trains and fine-tunes the RE model using consistent objectives based on contrastive learning. Contrastive learning often results in one relation forming multiple clusters in the representation space. The authors introduce a multi-center contrastive loss that facilitates the formation of various clusters for a single relation during pretraining, aligning it more effectively with the subsequent fine-tuning process. Notably, PubmedBERT and BERT-base-uncased are the chosen encoders for pre-training and fine-tuning. 

Finally, \cite{xu-etal-2023-s2ynre} introduces S2ynRE, a system designed to address the challenge of low-resource RE by harnessing the power of LLMs such as GPT-2 and GPT-3.5. These LLMs generate synthetic data and adapt to the target domain, automatically producing substantial amounts of coherent and realistic training data. The proposed algorithm employs a two-stage self-training approach, iteratively and alternately learning from synthetic and golden data. The effectiveness of the system is demonstrated across five diverse datasets. The paper uses BERT for baseline comparison and encoding, showcasing that combining large language models surpasses the current state-of-the-art in relation extraction. 

\subsubsection{Distant supervision Relation Extraction}\hfill\\


Distant supervision for RE provides a scalable and efficient way to generate large amounts of labeled training data, which is often scarce and expensive to obtain manually. In traditional supervised learning, high-quality labeled data is required for training models. However, manually annotating such data for RE tasks is labor-intensive, time-consuming, and costly, particularly for large and diverse datasets. Distant supervision automates this process by aligning existing structured data from knowledge bases with unstructured text, generating training examples without human intervention. Distant supervision techniques often generate training data by automatically aligning a knowledge base with unstructured text, which can lead to inaccuracies. These inaccuracies arise in the form of false positives (incorrectly labeled data) and false negatives (missing relations) due to the incompleteness of the knowledge base. Noisy data can significantly degrade model performance by introducing spurious patterns and incorrect information during training, leading to poor generalization and reduced accuracy in real-world applications. By effectively filtering out or mitigating the impact of noisy labels, models can better capture true relational patterns, enhance robustness, and improve their ability to extract relations from text accurately. This process is critical in RE tasks where precise and reliable identification of relationships is critical for downstream applications. Based on this motivation, most papers involving distant supervision for relation extraction focus on the necessity of techniques to address noisy data and mitigate the impact of negative examples.

Distant supervision often leads to noisy and incomplete labels, resulting in false negatives. In \cite{hao-etal-2021-knowing}, the authors addressed the problem of false negatives in distant supervision for relation extraction.  The proposed two-stage approach leverages the memory mechanism of deep neural networks to filter out noisy samples. It utilizes adversarial training to align unlabeled data with training data and generate pseudo labels for additional supervision. The approach achieves state-of-the-art results on two benchmark datasets, NYT10 and GIDS, outperforming several baseline models, including BERT. 

 \cite{sui-etal-2021-distantly-supervised} addresses the issue of label noise in distantly supervised relation extraction, which often results in inaccurate or incomplete outcomes. To mitigate this challenge, the authors introduce a denoising method based on multiple-instance learning. This method involves using multiple platforms to identify reliable sentences collectively. It is implemented within a federated learning framework, ensuring data decentralization and privacy protection by separating model training from direct access to raw data. In this study, BERT and RoBERTa are used as the encoders.

In the same line, Xie et al. \cite{xie-etal-2021-revisiting} focus on the problem of missing relations caused by the incompleteness of the knowledge base (false negatives) rather than reducing wrongly labeled relations (false positives). To address this issue, they propose a pipeline approach called RERE using BERT for encoding, which first performs sentence classification with relational labels and then extracts the subjects/objects. Furthermore, the proposed method, RERE, consistently outperforms existing approaches over the NYT dataset, even when learned with many false positive samples. However, the authors also experimented with a Chinese dataset (ACE05), using RoBERTa for encoding, but achieved poor performance. \cite{sun-etal-2023-uncertainty} introduces a framework, UG-DRE, which harnesses uncertainty estimation technology to guide the denoising of pseudo labels in distant supervision data. The proposal incorporates an instance-level uncertainty estimation method, gauging the reliability of pseudo labels with overlapping relations. Dynamic uncertainty thresholds are introduced for different types of ties to filter high-uncertainty pseudo labels. The authors employ BERT-base-uncased to capture contextual representations of documents and entities, facilitating the generation of pseudo labels and the denoising process. Results illustrate that using the discussed stages of uncertainty estimation, dynamic uncertainty threshold creation, and the iterative e-labeling strategy, the UG-DRE framework effectively mitigates noise in distant supervision data, enhancing performance in document-level RE tasks.


In addition to research addressing noisy data, we also found a line of work focused on enhancing distant supervision for relation extraction by incorporating external knowledge, fine-grained entity types, and innovative training strategies.

In \cite{christopoulou-etal-2021-distantly}, the authors introduced a multi-task probabilistic approach for distantly supervised RE, employing a variational autoencoder. The paper's primary contribution lies in integrating Knowledge Base priors into the variational autoencoder framework to enhance sentence space alignment and foster interpretability. To assess the proposed approach's effectiveness, the authors evaluated two benchmark datasets, NYT10 and WikiDistant. As an encoder, the authors leveraged a BILSTM model in conjunction with 50-dimensional embeddings provided by GLOVE.

In \cite{dai2021two}, the authors propose to leverage a Universal Graph (UG) that contains external knowledge to provide additional evidence for relation extraction. They introduce two training strategies: Path Type Adaptive Pretraining, which enhances the model's adaptability to various UG paths by pretraining on diverse path types, and Complexity Ranking Guided Attention, which enables the model to learn from both simple and complex UG paths by ranking their relevance and guiding attention accordingly. The paper evaluates the proposed framework on two datasets: a biomedical dataset (crafted by the authors using data from the Unified Medical Language System and Medline) and the NYT10 dataset. It shows that the proposed methods outperform several baseline methods.

In contrast to prior research, \cite{hu-etal-2021-semi-supervised} challenges the assumption of distant supervision by proposing that the relation between entity pairs should be context-independent. This assumption introduces context-agnostic label noises, leading to a phenomenon known as gradual drift. Addressing this challenge, the authors propose MetaSRE, a method that leverages unlabeled data to enhance the accuracy and robustness of RE by generating quality assessments on pseudo labels. MetaSRE employs two networks: the Relation Classification Network (RCN) and the Relation Label Generation Network (RLGN). The RCN is trained on labeled data, while the RLGN generates high-quality pseudo labels from unlabeled data. Subsequently, these pseudo labels train a meta-learner that can adapt to new domains and tasks. Experimental evaluations conducted on two public datasets, SemEval 2010 Task 8 and FewRel, demonstrated that MetaSRE surpassed other state-of-the-art methods in accuracy and robustness. Authors fine-tuned BERT on labeled data, utilizing it to generate features for the RCN. BERT was employed to generate embeddings for the unlabeled data, which the RLGN then used to create pseudo labels.

In \cite{shahbazi-etal-2020-relation}, Shahbazi presents a system that transforms entity mentions into fine-grained entity types (FGET), enhancing precision and explainability. Notably, it draws on typical relations from the FB-NYT dataset \cite{riedel2010modeling}, such as ``place lived'' and ``capital''. This paper relies on traditional word embeddings like skip-gram, showcasing their enduring efficacy in relation extraction. 

Finally, \cite{ma-etal-2021-sent} proposes an approach for sentence-level distant RE using negative training (NT) and presents a framework called SENT that filters noisy data and improves performance for downstream applications. The approach is based on the idea that selecting complementary labels of the given label during training decreases the risk of providing noisy information and prevents the model from overfitting the noisy data. The model trained with NT can separate the noisy data from the training data, significantly protecting the model from noisy information. The paper uses BERT to encode the input sentences and achieve state-of-the-art performance on TACRED and NYT-10 datasets.

\subsubsection{Open Relation Extraction}\hfill\\

OpenRE has seen significant advancements with approaches designed to address limitations of traditional clustering-based methods and enhance relation discovery in open settings. The problem with clustering-based methods is that they may need help to capture the advanced contextual information stored in vectors, leading to similar relations with different relationships in the same cluster.  In \cite{zhao-etal-2021-relation}, Zhao et al. leverage clustering methods and propose OpenRE, a method that addresses existing clustering-based approaches' limitations. Specifically, the authors propose a relation-oriented clustering method that leverages labeled data to learn a relation-oriented representation. The technique minimizes the distance between instances with the same relation and gathers instances towards their corresponding relation centroids to form the cluster structure. The authors demonstrate their method's effectiveness on two datasets, FewRel and TACRED, achieving a 29.2\% and 15.7\% reduction in error rate compared to current state-of-the-art methods. The proposed method also leverages BERT as the encoder.  

\cite{zhao-etal-2023-actively} introduces actively supervised clustering for Open Relation Extraction. This approach involves alternating between clustering learning and relation labeling. The aim is to furnish essential guidance for clustering while minimizing the need for additional human effort. The paper proposes active labeling strategy to discover clusters associated with unknown relations dynamically. In their experiments, the authors use TACRED and FewRel. Notably, they also incorporate the FewRel-LT dataset, an extended version of FewRel featuring a more diverse distribution of relations, particularly in the long tail — the paper references using BERT as an encoder. As a follow-up, \cite{zhao-etal-2023-open} introduce a method that tackles the issue of encountering unknown relations within the test set. Using BERT as an encoder to recognize known relations, the authors introduced a new relation, NOTA (none-of-the-above), to capture and understand previously unlearned relations.

\subsubsection{Multi-Task}\hfill\\

Most of the studies related to multi-task RE address the challenges of relation extraction at either the document or sentence level, particularly in low-resource settings. Most propose techniques to bridge the gaps in data availability and improve performance in these challenging scenarios. We also identify techniques that are effective across different levels of granularity, such as document-level and sentence-level relation extraction.

\cite{tian-etal-2021-dependency} incorporate dependency-type information to improve relation extraction further. By including the syntactic instruction among connected words, the proposed Attentive Graph Neural Network (A-GCN) model can better distinguish the importance of different word dependencies and improve the accuracy of relation extraction. The authors introduce type information into the A-GCN modeling to incorporate this information, effectively enhancing the model's performance. Furthermore, they used BERT as the model's encoder and introduced unique tokens to annotate the entities in the text, further improving the model's performance. Overall, their approach demonstrates the effectiveness of leveraging dependency trees and dependency types for relation extraction and achieves state-of-the-art performance on SemEval-2010 task8 \cite{hendrickx2019semeval} and ACE05 \cite{walker2006ace} datasets. 

The traditional feature extraction methods often extract features sequentially or in parallel. This process can result in suboptimal feature representation learning and limited task interaction. In the domain of joint models, Yan et al. \cite{yan-etal-2021-partition} introduces an approach to joint entity and RE utilizing a partition filter network.  Yan et al.'s partition filter network addresses this issue by generating task-specific features, enabling more effective two-way interaction between tasks. This approach ensures that features extracted later maintain direct contact with those extracted earlier. The authors conducted extensive experiments across six datasets, demonstrating that their method outperformed other baseline approaches. BERT-base-cased, ALBERTxxlarge-v1, and sciBERT-sci-vocab-uncased were employed as encoders for NER and RE partitions, but the main component in the classification stage is an LSTM.

\cite{eyal-etal-2021-bootstrapping} confronts the persistent challenge of acquiring a substantial volume of training data for relation extraction. The authors introduce a bootstrapping process for generating training datasets for relation extraction, specifically designed to be swiftly executed even by individuals without expertise in NLP. Their approach leverages search engines across syntactic graphs to procure positive examples, identifying sentences syntactically akin to user-input examples. This method is applied to relations sourced from TACRED and DocRED, demonstrating that the resultant models exhibit competitiveness with those trained on manually annotated data and data obtained through distant supervision. To further enhance the dataset, the authors employ pre-trained language models like BERT and RoBERTa for data augmentation. This strategy involves generating additional relation examples using GPT-2 and manual annotation. In particular, the authors modify the text by substituting the entity pair with unique tokens, feeding the altered text as input to a BERT model. The relation between the two entities is captured by concatenating the final hidden states corresponding to their respective start tokens. This representation undergoes classification through a dedicated head, and the model is fine-tuned for relation classification. 

A notable challenge in RE lies in the variation of relation types across different datasets, complicating efforts to amalgamate them for training a unified model. To tackle this issue, \cite{fu-grishman-2021-learning} propose a method that employs prototypes of relation types to discern their relatedness within a multi-task learning framework. These prototypes are constructed by selecting representative examples of each relation type and using them to augment related types from a distinct dataset. The authors conduct experiments using two datasets featuring similar relation schemas: the ACE05 and ERE dataset \cite{aguilar2014comparison}. While the paper mentions using an encoder, it does not specify the language model leveraged in the experiments.

Continuing with joint methodologies, Zheng et al. \cite{zheng-etal-2023-jointprop} introduce Jointprop, a comprehensive graph-based framework designed for joint semi-supervised entity and relation extraction, specifically tailored to address the challenges posed by limited labeled data. By leveraging heterogeneous graph-based propagation, Jointprop adeptly captures global structural information among individual tasks, allowing for a more integrated approach to understanding the relationships between entities and their corresponding relations. This framework facilitates the propagation of labels across a unified graph of labeled and unlabeled data and exploits interactions within the unlabeled data to enhance learning. Experimental results underscore the effectiveness of this approach, producing state-of-the-art performance on renowned datasets, including SciERC, ACE05, ConLL, and SemEval, with notable improvements in F1 scores for both entity recognition and relation extraction tasks. The paper discusses using language models to obtain contextualized representations for each token. Although the specific language model used is not mentioned, the paper suggests that BERT could be a potential model for creating these representations. This implies that Jointprop could gain from the detailed contextual embeddings offered by transformer-based models, thus improving its performance in settings with limited resources.

\subsection{Observations}

Regarding models, we identified four main groups: those employing LSTMs, those utilizing CNNs and graph-based approaches, and the largest group leveraging transformers like BERT. It is important to note that many works incorporate language models like BERT at some stage for encoding, with classification often performed using other techniques, such as LSTMs. However, in this section, we categorize each work based on its primary component. The presence or absence of BERT as an encoder or as part of the main RE component is detailed in Table \ref{tab:model-usage}.

Additionally, we observed that not all works adhere to the entire three-stage RE pipeline, with some focusing solely on entity detection or classifying pre-tagged entities within a dataset. To present our findings concisely, we have organized all reviewed works into tables categorized by conference and year, offering a comprehensive year-conference-task perspective on the cutting-edge developments in RE. It is also important to note that many reviewed papers do not explicitly specify how many stages of the relation extraction process they perform. We have inferred this information based on their models, methodologies, and reported results in such cases. Table \ref{tab:model-usage-acl} presents a summary of insights from the ACL conference, while Table \ref{tab:model-usage-others} covers insights from the ACL satellite conferences. Finally, Table \ref{tab:summary} offers a comprehensive overview of the number of papers that utilized each model or addressed each task. This summary allows us to derive key insights into trends over the years.

\begin{table}[htb]
\centering
 \small
    \begin{adjustbox}{max width=\textwidth}
\begin{tabular}{c c c c c c}
\hline
\textbf{Paper} & \textbf{Conference} & \textbf{Year} & \textbf{Model Group} & \textbf{Task / Challenge} & \textbf{Subtask} \\
\hline
\cite{shahbazi-etal-2020-relation} & ACL & 2020 & CNN & Distant Supervised RE & Relation Classification \\
\cite{nan-etal-2020-reasoning} & ACL & 2020 & Graph Based & Document Level & Relation Classification \\
\cite{yu-etal-2020-dialogue} & ACL & 2020 & Transformer & Sentence Level & Relation Identification + Relation Classification \\
\cite{tian-etal-2021-dependency} & ACL & 2021 & CNN & Sentence Level / Multi-task & Relation Classification \\
\cite{ma-etal-2021-sent} & ACL & 2021 & LSTM & Distant Supervised RE & Relation Identification + Relation Classification \\
\cite{huang-etal-2021-three} & ACL & 2021 & LSTM & Document Level & Relation Classification \\
\cite{xie-etal-2021-revisiting} & ACL & 2021 & Transformer & Distant Supervised RE & NER + Relation Classification \\
\cite{yang-etal-2021-entity} & ACL & 2021 & Transformer & Few Shot / Low Resource & Relation Classification \\
\cite{huang-etal-2021-entity} & ACL & 2021 & Transformer & Document Level & Relation Classification \\
\cite{wang-etal-2021-unire} & ACL & 2021 & Transformer & Sentence Level & NER + Relation Identification + Relation Classification \\
\cite{qin-joty-2022-continual} & ACL & 2022 & Transformer & Few Shot / Low Resource & Relation Classification \\
\cite{liu-etal-2022-pre} & ACL & 2022 & Transformer & Few Shot / Low Resource & Relation Classification \\
\cite{jie-etal-2022-learning} & ACL & 2022 & Transformer & Sentence Level & NER (quantities) + Relation Classification \\
\cite{gururaja-etal-2023-linguistic} & ACL & 2023 & Graph Based & Few Shot / Low Resource & Relation Classification \\
\cite{zheng-etal-2023-jointprop} & ACL & 2023 & Graph Based & Multi-Task & NER + Relation Classification \\
\cite{wu-etal-2023-information} & ACL & 2023 & Graph Based & Multilingual and multimodal Relation Extraction & NER + Relation Identification + Relation Classification \\
\cite{xu-etal-2023-nli} & ACL & 2023 & Transformer & Few Shot / Low Resource & Relation Classification \\
\cite{hennig-etal-2023-multitacred} & ACL & 2023 & Transformer & Multilingual and multimodal Relation Extraction & Relation Classification \\
\cite{huguet-cabot-etal-2023-red} & ACL & 2023 & Transformer & Multilingual and multimodal Relation Extraction & NER + Relation Classification \\
\cite{zhao-etal-2023-actively} & ACL & 2023 & Transformer & Open RE & Relation Classification \\
\cite{zhao-etal-2023-actively} & ACL & 2023 & Transformer & Document Level & Relation Classification \\
\cite{zhao-etal-2023-matching} & ACL & 2023 & Transformer & Few Shot / Low Resource & Relation Identification + Relation Classification \\
\cite{zheng-etal-2023-rethinking} & ACL & 2023 & Transformer & Multilingual and multimodal Relation Extraction & NER + Relation Classification \\
\cite{xu-etal-2023-s2ynre} & ACL & 2023 & Transformer & Few Shot / Low Resource & NER + Relation Identification + Relation Classification \\
\cite{zhao-etal-2023-open} & ACL & 2023 & Transformer & Open RE & Relation Identification + Relation Classification \\
\cite{zhang-etal-2023-novel} & ACL & 2023 & Transformer & Document Level & NER + Relation Classification \\
\cite{zhou-etal-2023-continual} & ACL & 2023 & Transformer & Few Shot / Low Resource & NER + Relation Classification \\
\cite{sun-etal-2023-uncertainty} & ACL & 2023 & Transformer & Distant Supervised RE & Relation Classification \\
\cite{yuan-etal-2023-discriminative} & ACL & 2023 & Transformer & Document Level & Relation Classification \\
\cite{hu-etal-2023-multimodal} & ACL & 2023 & Transformer & Multilingual and multimodal Relation Extraction & NER + Relation Identification + Relation Classification \\
\cite{zhao-etal-2023-improving} & ACL & 2023 & Transformer & Document Level & Relation Classification \\
\cite{yang-etal-2023-histred} & ACL & 2023 & Transformer & Multilingual and multimodal Relation Extraction & Relation Classification \\
\hline
\end{tabular}
\end{adjustbox}
\caption{Summary of key papers in ACL categorized by model group and task}
\label{tab:model-usage-acl}
\end{table}

\begin{table}[hbt]
\centering
 \small
    \begin{adjustbox}{max width=\textwidth}
\begin{tabular}{c c c c c c}
\hline
\textbf{Paper} & \textbf{Conference} & \textbf{Year} & \textbf{Model Group} & \textbf{Task / Challenge} & \textbf{Subtask} \\
\hline
\cite{yuan-eldardiry-2021-unsupervised} & EACL & 2021 & CNN & Sentence Level & Relation Classification \\
\cite{hao-etal-2021-knowing} & EACL & 2021 & CNN & Distant Supervised RE & Relation Identification + Relation Classification \\
\cite{sui-etal-2021-distantly-supervised} & EACL & 2021 & CNN & Distant Supervised RE & Relation Classification \\
\cite{zhang-etal-2021-entity} & EACL & 2021 & LSTM & Document Level & NER + Relation Classification \\
\cite{yan-etal-2021-partition} & EACL & 2021 & LSTM & Multi-Task & NER + Relation Classification \\
\cite{brody-etal-2021-towards} & EACL & 2021 & Transformer & Few Shot / Low Resource & Relation Classification \\
\cite{eberts-ulges-2021-end} & EACL & 2021 & Transformer & Document Level & NER + Relation Identification + Relation Classification \\
\cite{zhao-etal-2021-relation} & EACL & 2021 & Transformer & Open RE & Relation Classification \\
\cite{eyal-etal-2021-bootstrapping} & EACL & 2021 & Transformer & Multi-Task & Relation Classification \\
\cite{fu-grishman-2021-learning} & EACL & 2021 & Transformer & Multi-Task & NER + Relation Identification + Relation Classification \\
\cite{han-etal-2021-exploring} & EACL & 2021 & Transformer & Few Shot / Low Resource & Relation Classification \\
\cite{hu-etal-2021-semi-supervised} & EACL & 2021 & Transformer & Distant Supervised RE & Relation Classification \\
\cite{hu-etal-2021-gradient} & EACL & 2021 & Transformer & Few Shot / Low Resource & Relation Classification \\
\cite{seganti-etal-2021-multilingual} & EACL & 2021 & Transformer & Multilingual and multimodal Relation Extraction & NER + Relation Classification \\
\cite{christopoulou-etal-2021-distantly} & NAACL & 2021 & LSTM & Distant Supervised RE & NER + Relation Identification + Relation Classification \\
\cite{zhong-chen-2021-frustratingly} & NAACL & 2021 & Transformer & Sentence Level & NER + Relation Identification + Relation Classification \\
\cite{chen-li-2021-zs} & NAACL & 2021 & Transformer & Few Shot / Low Resource & Relation Classification \\
\cite{yang-song-2022-fpc} & AACL & 2022 & Transformer & Sentence Level & Relation Classification \\
\cite{yan-etal-2022-empirical} & AACL & 2022 & Transformer & Sentence Level & NER + Relation Classification \\
\cite{arcan-etal-2022-towards} & AACL & 2022 & Transformer & Few Shot / Low Resource & NER + Relation Identification + Relation Classification \\
\cite{zhou-chen-2022-improved} & AACL & 2022 & Transformer & Sentence Level & NER + Relation Classification \\
\cite{xu-etal-2022-document} & NAACL & 2022 & Graph Based & Document Level & Relation Classification \\
\cite{wang-etal-2022-rely} & NAACL & 2022 & Graph Based & Sentence Level & NER + Relation Classification \\
\cite{xu-choi-2022-modeling} & NAACL & 2022 & Graph Based & Document Level & NER + Relation Classification \\
\cite{tiktinsky-etal-2022-dataset} & NAACL & 2022 & Transformer & Few Shot / Low Resource & NER + Relation Identification + Relation Classification \\
\cite{liang-etal-2022-modeling} & NAACL & 2022 & Transformer & Sentence Level & Relation Classification \\
\cite{popovic-farber-2022-shot} & NAACL & 2022 & Transformer & Few Shot / Low Resource & Relation Classification \\
\cite{liu-etal-2022-hiure} & NAACL & 2022 & Transformer & Sentence Level & Relation Classification \\
\cite{xiao-etal-2022-sais} & NAACL & 2022 & Transformer & Document Level & NER + Relation Classification \\
\cite{ma-etal-2023-dreeam} & EACL & 2023 & Transformer & Document Level & Relation Identification + Relation Classification \\
\cite{teru-2023-semi} & EACL & 2023 & Transformer & Few Shot / Low Resource & Relation Classification \\
\cite{guo-etal-2023-towards} & EACL & 2023 & Transformer & Document Level & Relation Classification \\
\cite{najafi-fyshe-2023-weakly} & EACL & 2023 & Transformer & Few Shot / Low Resource & Relation Identification + Relation Classification \\
\hline

\end{tabular}
\end{adjustbox}
\caption{Summary of key papers in EACL, NAACL, and AACL categorized by model group and task}
\label{tab:model-usage-others}
\end{table}

\begin{table}[htb]
\centering
 \small
    \begin{adjustbox}{max width=\textwidth}
\begin{tabular}{l c c c c c c c c c c c}
\hline
\textbf{Conference} & \textbf{Year} & \textbf{CNN} & \textbf{LSTM} & \textbf{Transformer} & \textbf{Graph Based} & \textbf{Sentence Level} & \textbf{Document Level} & \textbf{Multi-Task} & \textbf{Distant Supervised RE} & \textbf{Few Shot / Low Resource} & \textbf{Multilingual and Multimodal} \\
\hline
ACL & 2020 & 1 & 0 & 1 & 1 & 1 & 0 & 0 & 1 & 0 & 0 \\
ACL & 2021 & 2 & 2 & 5 & 0 & 1 & 2 & 1 & 3 & 1 & 1 \\
ACL & 2022 & 0 & 0 & 7 & 1 & 1 & 0 & 0 & 0 & 3 & 0 \\
ACL & 2023 & 0 & 0 & 14 & 1 & 3 & 3 & 1 & 1 & 6 & 5 \\
EACL & 2021 & 3 & 2 & 9 & 0 & 2 & 2 & 2 & 3 & 3 & 1 \\
NAACL & 2021 & 0 & 1 & 2 & 0 & 1 & 0 & 0 & 1 & 1 & 0 \\
AACL & 2022 & 0 & 0 & 5 & 0 & 3 & 0 & 0 & 0 & 2 & 0 \\
NAACL & 2022 & 0 & 0 & 5 & 3 & 3 & 3 & 0 & 0 & 2 & 0 \\
EACL & 2023 & 0 & 0 & 6 & 0 & 0 & 3 & 0 & 0 & 2 & 0 \\
\hline
\end{tabular}
\end{adjustbox}
\caption{Summary of models and tasks by conference and year.}
\label{tab:summary}
\end{table}

One of the key insights from the trend analysis is related to the models used. In the first two years of conferences, 2020 and 2022, we observe a trend toward using various models, including CNNs and LSTMs, though they are less prevalent. By 2022 and 2023, however, the focus has shifted to transformer-based techniques, highlighting their growing dominance and effectiveness. 

When examining the subtasks, it is clear that relation identification is the least frequently addressed. Most datasets are tagged with pre-identified relations and are used primarily for relation classification. This subtask, the primary focus in the literature, remains the most relevant component of relation extraction techniques.

Examining the tasks and challenges within relation extraction reveals some compelling insights. Specifically, document-level relation extraction has gained significant prominence in the past two years, mainly due to the rise of transformer-based techniques capable of processing more intricate and extensive contexts. Data show that 54\% of papers addressing this issue utilize transformer architecture, supporting this trend.

The trend is even more pronounced in few-shot and low-resource scenarios, where the generalizability of language models offers the best solutions. Nearly 100\% of the papers in these areas employ transformer-based approaches, reflecting a broader shift towards these techniques in recent years. This shift is likely tied to the widespread adoption and democratization of language models.

These observations are further supported by our findings in Tables \ref{tab:docre_results} and \ref{tab:zsre_results}, highlighting the best models' performance in these domains.

\section{Large Language Models vs language models in RE}
\label{sec:llmvslm}

We reviewed various baseline approaches in the literature and compared the effectiveness of different language models for encoding relations. We survey and analyze existing studies that use various language models, including GPT-3, T5, and traditional models like BERT or RoBERTa. Our aim is two-fold: firstly, to assess the relative performance of different model categories in the challenging task of RE, and secondly, to evaluate whether the emergence of Large Language Models represents a significant advancement over traditional models in performance. Our study delves into Few-Shot relation extraction, document-level relation extraction, and traditional relation extraction. We have selected the top five models for each benchmark dataset associated with these tasks: FewRel, DocRed, and TACRED. This approach allows us to capture various methodologies and performances across distinct facets of RE challenges. Table \ref{tab:re_results} presents the results for TACRED. For document-level relation extraction on the DocRED dataset, the outcomes are detailed in Table \ref{tab:docre_results} and Table \ref{tab:zsre_results} compiles the results for few-shot relation extraction using the FewRel dataset.

Examining the aggregated results reveals a clear dominance of BERT-based approaches, with BERT and RoBERTa collectively representing a substantial 75\% of the primary outcomes. In contrast, Large Language Models (LLMs) contribute merely 25\% to the overall results. A temporal analysis unveils intriguing trends, particularly in the latest year, 2023. This year has proven pivotal, delivering state-of-the-art results in two out of the three experiments. Notably, BERT and RoBERTa continue to secure the top positions, while other LLMs, such as T5, claim third place, highlighting the sustained prominence of BERT-based models.

\begin{table}[htb]
\centering
\begin{tabular}{l c c c c}
\hline
\textbf{Reference} & \textbf{Year} & \textbf{Model} & \textbf{F1} & \textbf{LM/LLM} \\
\hline
Wang et al. (2022) \cite{wang-etal-2022-deepstruct} & 2022 & DeepStruct & 76.8 & GLM \\
Huang et al. (2022) \cite{huang-etal-2022-unified} & 2022 & UNIST & 75.5 & RoBERTa \\
Baek et al. (2022) \cite{baek2022enhancing} & 2022 & RE-MC & 75.4 & RoBERTa \\
Han et al. (2022) \cite{han-etal-2022-generative} & 2022 & GEN-PT & 75.3 & T5 \\
Lyu and Chen (2021) \cite{lyu-chen-2021-relation} & 2021 & BERT & 75.2 & BERT \\
\hline
\end{tabular}
\caption{State-of-the-art results for sentence-level RE using the TACRED dataset}
\label{tab:re_results}
\end{table}

\begin{table}[htb]
\centering
\begin{tabular}{l c c c c}
\hline
\textbf{Reference} & \textbf{Year} & \textbf{Model} & \textbf{F1} & \textbf{LM/LLM} \\
\hline
Ma et al. (2023) & 2023 & DREEAM & 67.53 & RoBERTa \\
Tan et al. (2022) & 2022 & KD-Rb-l & 67.28 & RoBERTa \\
Xu et al. (2021) & 2021 & SSAN-RoBERTa-large & 65.92 & RoBERTa \\
Xiao et al. (2022) & 2022 & SAIS-RoBERTa-large & 65.11 & RoBERTa \\
Xie et al. (2022) & 2022 & Eider-RoBERTa-large & 64.79 & RoBERTa \\
\hline
\end{tabular}
\caption{State-of-the-art results for Document-Level RE in the DocRED dataset}
\label{tab:docre_results}
\end{table}

In the context of the benchmark dataset DocRED (Table \ref{tab:docre_results}), the top-performing models consistently employ RoBERTa large as their encoder, a noteworthy observation with implications for addressing our RQ4. It is striking that there is a conspicuous absence of papers or experiments using language models other than RoBERTa or BERT.

Large language models like RoBERTa and BERT are widely used and effectively extract relationships between different document elements. This is also true for extracting relationships at the sentence level. The reason for this effectiveness is linked to the Masked Language Model (MLM) strategy. This strategy involves training the model on a large body of text by hiding a word within the context of other words, similar to how the model processes relation classification tasks, which involve identifying relationships between two entities.

Upon deeper scrutiny, we find that RoBERTa's consistent prominence in top positions for document-level relation extraction is linked to intrinsic details of the model:

\begin{itemize}
    \item RoBERTa uses a much larger dataset for pretraining, encompassing more data and longer sequences. This extensive training regime contributes to the model's robustness and generalization.
    \item RoBERTa employs only the MLM objective during pretraining, discarding the Next Sentence Prediction (NSP) objective. Sentences are treated as continuous streams of text without explicit sentence separation, making it particularly suitable for the challenges posed by document-level relation extraction.
    \item RoBERTa allows training with longer sequences, surpassing the 512-token limit BERT uses. This capability provides a more comprehensive and extended context, making it well-suited for document-level relation extraction tasks.
\end{itemize}

\begin{table}[htb]
\centering
\begin{tabular}{l c c c c}
\hline
\textbf{Reference} & \textbf{Year} & \textbf{Model} & \textbf{F1} & \textbf{LM/LLM} \\
\hline
Tran et al. (2023) & 2023 & ESC-ZSRE & 81.68 & BERT \\
Chia et al. (2022) & 2022 & RelationPrompt & 79.96 & GPT2+BART \\
Tran et al. (2022) & 2022 & IDL & 62.61 & BERT \\
Najafi and Fyshe (2023) & 2023 & OffMML-G(+negs) & 61.3 & T5 \\
Chen and Li (2021) & 2021 & ZS-BERT & 57.25 & BERT \\
\hline
\end{tabular}
\caption{State-of-the-art results for Zero-Shot RE models in the FewRel dataset}
\label{tab:zsre_results}
\end{table}

Large Language Models (LLMs) excel in few-shot learning scenarios. This proficiency is attributed to their extensive pretraining on massive volumes of textual data, enabling them to grasp diverse linguistic patterns and acquire broad knowledge. The efficacy of this pretraining becomes evident as LLMs demonstrate a remarkable ability to generalize effectively to new tasks with only a limited number of examples, positioning them as ideal candidates for few-shot learning applications. We identify substantial potential for further exploration and research involving LLMs in this domain. 

\section{Discussion}
\label{sec:discusion}

In our descriptive analysis, w have noticed a growing interest in using language models to solve Relation Extraction (RE) problems. As is shown in Figure 2, the number of publications on this topic has been consistently increasing, with the most recent year reaching the highest point. This continued growth shows that the field is active and encourages more research and development.

\begin{figure}[hbt]
  \centering
  \includegraphics[width=0.5\linewidth]{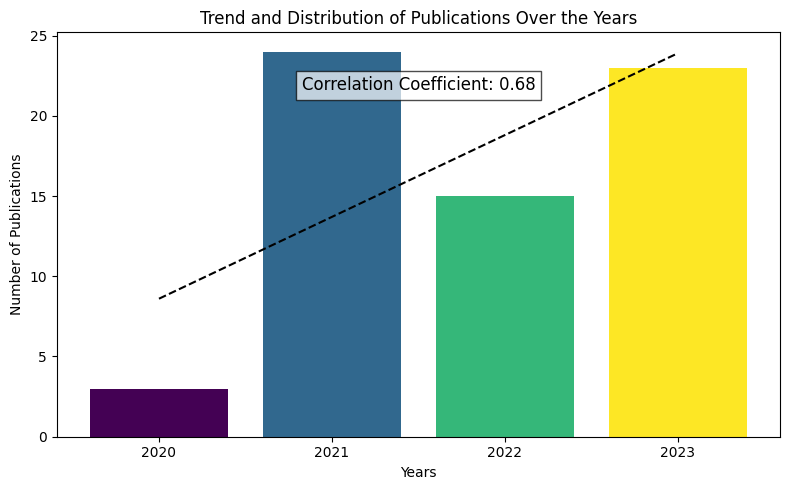}
  \caption{Trend and Distribution of Publications Over the Years}
  \label{fig:publications}
\end{figure}

\begin{figure}[hbt]
  \centering
  \includegraphics[width=0.5\linewidth]{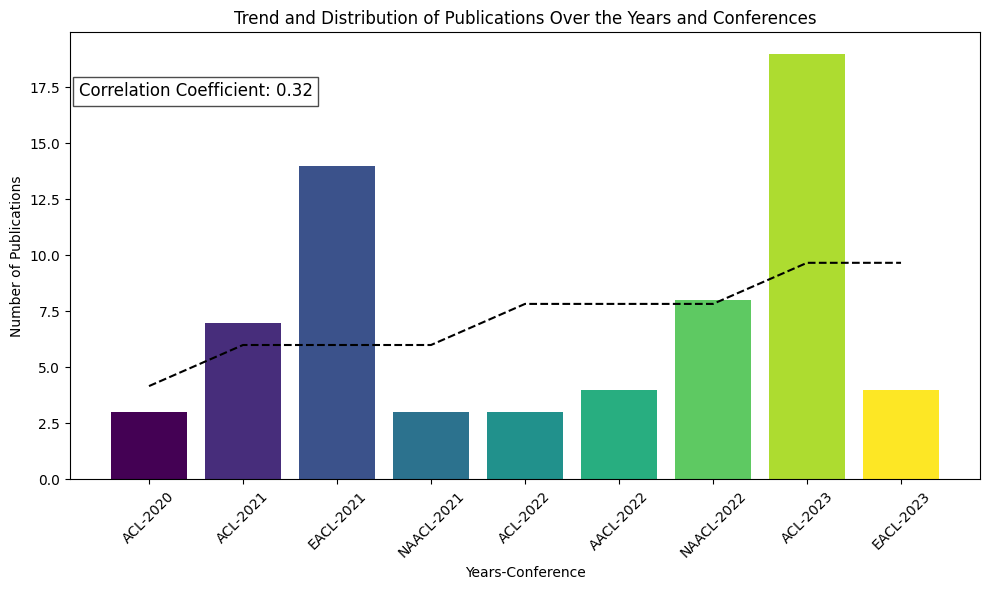}
  \caption{Trend and Distribution of Publications Over the Years and conferences}
  \label{fig:publications-2}
\end{figure}

It is important to consider that some conferences occur biennially. We created a year-conference graph in Figure \ref{fig:publications-2} to address this. Although the data for 2023 appears underrepresented compared to 2022, with 3 and 2 conferences, the overall trend remains upward. The graph in Figure \ref{fig:publications} shows that the higher number of publications in 2021 is mainly due to more conferences than in 2023.

We calculated the correlation coefficients for both graphs for a more comprehensive analysis. The cumulative graph by year shows a correlation coefficient of 0.68, while the year-conference graph shows a coefficient of 0.32. These values suggest an increase as the years progress.

In terms of application areas, our analysis reveals that relation extraction has found application in diverse domains, including media, academia, economics, and even unconventional realms such as heritage conservation and mathematics. This broad applicability underscores the technique's usefulness and emphasizes the need for new systems capable of autonomously identifying novel relations in an ever-changing landscape. 

In the upcoming subsections, we lay the groundwork for addressing current challenges while providing an exhaustive review of language models used for RE. Our approach is to align our findings with our RQ in a concerted effort to contribute meaningfully to ongoing research in the field of RE.

\subsection{RQ1: What are the challenges of RE that are being solved by systems that leverage language models?}

We have identified areas for improvement and further exploration in the following challenges:

\begin{itemize}
    \item \textbf{Document-level RE}:  While progress has been made in sentence-level RE, document-level RE remains challenging. Extending the capabilities of language models to capture relationships and context across entire documents deserves further attention. With their ability to capture broader contexts, large language models present a promising avenue for exploration in this area.

    \item \textbf{Multimodal RE}: Integrating information from diverse modalities, such as text and images, poses a distinct challenge. Enhancing language models to extract relations from multimodal data effectively requires fusing models for text and pictures. Despite some progress, there is still room for improvement, especially in benchmark datasets.
    
    \item \textbf{Multilingual RE}: Handling relations in languages not encountered during training or extracting relations with no training examples remains challenging. While many language models can generalize to other languages, their performance may still need to be bettered. Developing systems and training strategies, especially in transfer learning, is essential to address this challenge effectively.
    
    \item \textbf{Few-Shot and Low-Resource Relation Extraction}: This category encompasses areas with a scarcity of training examples, such as medical or biological datasets, and relations in traditional domains but with unseen relations. Large language models can offer valuable solutions in these domains, leveraging their ability to generate new content for training strategies and learn complex relations and contexts. Significant improvements have been observed in distant supervision RE, where automatically labeled datasets are created from knowledge bases, and large language models can contribute to further enhancements.

    \item\textbf{Open Relation Extraction}: Developing techniques to identify and characterize relations without predefined categories is an area that requires further investigation. Unsupervised techniques, such as clustering and large language models, hold promise in addressing this challenge.
\end{itemize}

RE grapples with challenges related to ambiguity and polysemy, where entities and relations frequently exhibit multiple meanings across different contexts. LMs, with their contextual understanding, play a crucial role in capturing the nuanced nature of language and disambiguating entities and relations based on the surrounding context. However, there is room for improvement. A promising avenue for further research is enhancing performance through the use of similarities or refining semantic searches. Finally, RE techniques must contend with negations and expressions of uncertainty in language. Ongoing research focuses on developing methods to handle the negation of relations or the presence of "none of the above" relations, reflecting the persistent challenges in dealing with negations and uncertain language expressions.

\subsection{RQ2: What are the most commonly used language models for the RE problem?}

Table \ref{tab:model-usage} illustrates the adoption of various language models and word embeddings for state-of-the-art relation extraction solutions in ACL conferences. It is worth mentioning that we have considered the base model if a paper does not explicitly specify the submodel used. Also, it is necessary to mention that we have condensed all the papers created over the survey period in this table. In this table, we have discussed the papers proposing a cutting-edge technique from section \ref{sec:review} and the papers proposing a dataset from section \ref{sec:datasets}. 

\begin{table}[htb]
    \centering
    \caption{Most widespread models}
    \label{tab:model-usage}
    \begin{tabular}{llcc}
        \toprule
        \textbf{Model} & \textbf{Sub-model} & \textbf{Times used} & \textbf{Papers that used the model} \\
        \midrule
       BERT & base & 45 & \parbox{8cm}{\centering\cite{zhao-etal-2023-improving,hennig-etal-2023-multitacred,zhao-etal-2023-actively,zhao-etal-2023-matching,zheng-etal-2023-rethinking,gururaja-etal-2023-linguistic,xu-etal-2023-s2ynre,zhao-etal-2023-open,zhang-etal-2023-novel,zhou-etal-2023-continual,sun-etal-2023-uncertainty,yuan-etal-2023-discriminative,hu-etal-2023-multimodal,ye-etal-2022-packed,qin-joty-2022-continual,jie-etal-2022-learning,bassignana-plank-2022-mean,wang-etal-2021-unire,xie-etal-2021-revisiting,tian-etal-2021-dependency,ma-etal-2021-sent,yang-etal-2021-entity,huang-etal-2021-entity,nan-etal-2020-reasoning,yu-etal-2020-dialogue,pouran-ben-veyseh-etal-2020-exploiting,brody-etal-2021-towards,eberts-ulges-2021-end,zhao-etal-2021-relation,eyal-etal-2021-bootstrapping,han-etal-2021-exploring,zhang-etal-2021-entity,yan-etal-2021-partition,hao-etal-2021-knowing,hu-etal-2021-semi-supervised,sui-etal-2021-distantly-supervised,hu-etal-2021-gradient,zhou-chen-2022-improved,yan-etal-2022-empirical,teru-2023-semi,zhong-chen-2021-frustratingly,xiao-etal-2022-sais,xu-etal-2022-document,liang-etal-2022-modeling,popovic-farber-2022-shot,liu-etal-2022-hiure}} \\
        RoBERTa & large & 10 & \parbox{8cm}{\centering\cite{zhang-etal-2023-novel,liu-etal-2022-pre,jie-etal-2022-learning,huang-etal-2021-entity,brody-etal-2021-towards,yang-song-2022-fpc,zhou-chen-2022-improved,ma-etal-2023-dreeam,guo-etal-2023-towards,xiao-etal-2022-sais}} \\
        SciBERT & scvocab & 7 & \parbox{8cm}{\centering\cite{ye-etal-2022-packed,bassignana-plank-2022-mean,yan-etal-2021-partition,yan-etal-2022-empirical,zhong-chen-2021-frustratingly,xiao-etal-2022-sais,tiktinsky-etal-2022-dataset}} \\
        GLOVE & & 4 & \parbox{8cm}{\centering\cite{huang-etal-2021-three,kruiper-etal-2020-laymans,nan-etal-2020-reasoning,christopoulou-etal-2021-distantly}} \\
        ALBERT & xxlarge & 4 & \parbox{8cm}{\centering\cite{ye-etal-2022-packed,wang-etal-2021-unire,yan-etal-2021-partition,zhong-chen-2021-frustratingly}} \\
        RoBERTa & base & 4 & \parbox{8cm}{\centering\cite{zhang-etal-2023-novel,brody-etal-2021-towards,eyal-etal-2021-bootstrapping,wang-etal-2022-rely}} \\
        CNN-based-encoder & & 3 & \parbox{8cm}{\centering\cite{dai-etal-2021-two,brody-etal-2021-towards,yuan-eldardiry-2021-unsupervised}} \\
        T5 & & 3 & \parbox{8cm}{\centering\cite{wadhwa-etal-2023-revisiting,arcan-etal-2022-towards,najafi-fyshe-2023-weakly}} \\
        BERT & large & 3 & \parbox{8cm}{\centering\cite{tian-etal-2021-dependency,zhou-chen-2022-improved,liang-etal-2022-modeling}} \\
        Span-BERT & & 3 & \parbox{8cm}{\centering\cite{brody-etal-2021-towards,liang-etal-2022-modeling,xu-choi-2022-modeling}} \\
        GPT & 3.5 & 2 & \parbox{8cm}{\centering\cite{xu-etal-2023-s2ynre,wadhwa-etal-2023-revisiting}} \\
        GPT & 2 & 2 & \parbox{8cm}{\centering\cite{xu-etal-2023-s2ynre,eyal-etal-2021-bootstrapping}} \\
        Word2Vec & & 2 & \parbox{8cm}{\centering\cite{pouran-ben-veyseh-etal-2020-exploiting,zhang-etal-2021-entity}} \\
        PubMedBERT & & 2 & \parbox{8cm}{\centering\cite{zhou-etal-2023-continual,tiktinsky-etal-2022-dataset}} \\
        Word2Vec &  Skip-Gram & 1 & \parbox{8cm}{\centering\cite{shahbazi-etal-2020-relation}} \\
        BioLinkBERT & large & 1 & \parbox{8cm}{\centering\cite{xu-etal-2023-nli}} \\
        AnchiBERT & & 1 & \parbox{8cm}{\centering\cite{yang-etal-2023-histred}} \\
        KLUE & & 1 & \parbox{8cm}{\centering\cite{yang-etal-2023-histred}} \\
        mBART & 50 & 1 & \parbox{8cm}{\centering\cite{huguet-cabot-etal-2023-red}} \\
        ViT & base & 1 & \parbox{8cm}{\centering\cite{wu-etal-2023-information}} \\
        BERT & Chinese & 1 & \parbox{8cm}{\centering\cite{jie-etal-2022-learning}} \\
        RoBERTa & Chinese & 1 & \parbox{8cm}{\centering\cite{jie-etal-2022-learning}} \\
        ELMO & & 1 & \parbox{8cm}{\centering\cite{kruiper-etal-2020-laymans}} \\
        LUKE & base & 1 & \parbox{8cm}{\centering\cite{brody-etal-2021-towards}} \\
        M-BERT & & 1 & \parbox{8cm}{\centering\cite{seganti-etal-2021-multilingual}} \\
        Sentence-BERT & & 1 & \parbox{8cm}{\centering\cite{chen-li-2021-zs}} \\
        BlueBERT & & 1 & \parbox{8cm}{\centering\cite{tiktinsky-etal-2022-dataset}} \\
        BioBERT & & 1 & \parbox{8cm}{\centering\cite{tiktinsky-etal-2022-dataset}} \\
        \bottomrule
    \end{tabular}
\end{table}

It is evident that BERT stands out as the predominant model across the literature, featuring prominently in 45 different papers, constituting almost 55\% of the recent papers about extraction. The second most widely used model is RoBERTa achieving state-of-the-art results, specially in areas like document-level relation extraction. When examining the percentage of papers that incorporate or at least mention large language models such as T5 or GPT in their methodology, we find that only 8.5\% of papers leverage these language models.

BERT is one of the oldest models studied in this survey. Its popularity may be influenced by the fact that it was one of the first language models widely adopted. To address this potential bias, we performed an analysis that normalizes the number of times a paper references a model by the number of years since model release. The formula used is $tu/y$, where $tu$ represents the frequency of use as shown in Table \ref{tab:model-usage}, and $y$ denotes the number of years from model release to the end of the survey in 2023. In case of BERT, was released in 2018 so, $y=2023-2018$, 5. This approach yields a factor of 9, which remains higher than newer models like T5 or GPT.  Furthermore, it is essential to consider potential issues on the use and adoption between BERT-based models and other large LLMs like GPT or T5. However, thanks to platforms like Huggingface's Transformers library \cite{wolf2019huggingface}, most models have been made widely available, making them widely accessible in their pre-trained forms. Given that researchers predominantly rely on transfer learning or fine-tuning rather than extensive training from scratch, we can conclude that access issues are minimal or nonexistent.

To provide a more accurate comparison, considering that BERT was released significantly earlier than other language models, we analyzed the data from the most recent year of the survey, 2023. By focusing on 2023, we ensure a fair evaluation of the models. We normalized the percentages and counts based on the release year of each model that had at least one citation in the 2023 survey papers. The results are presented in Table \ref{tab:model-usage-normalized}. 

\begin{table}[hbt]
\caption{2023 Papers by model variant with normalization by year}
 \label{tab:model-usage-normalized}
\centering
\small
\begin{adjustbox}{max width=\textwidth}
\begin{tabular}{llcccccc}
\hline
\textbf{Model} & \textbf{Variant} & \textbf{Reelease Year} & \textbf{Count} & \textbf{\%} & \textbf{Norm. Count} & \textbf{Norm. \%} & \textbf{References} \\ \hline
BERT & base & 2018 & 9 & 45\% & \textbf{1.80} & \textbf{11.54\%} & \parbox{8cm}{\centering\cite{zhao-etal-2023-improving,hennig-etal-2023-multitacred,zhao-etal-2023-actively,zhao-etal-2023-matching,zheng-etal-2023-rethinking,gururaja-etal-2023-linguistic,xu-etal-2023-s2ynre,zhao-etal-2023-open,zhang-etal-2023-novel}} \\ 
RoBERTa & large & 2019 & 2 & 10\% & \textbf{0.50} & \textbf{3.21\%} & \parbox{8cm}{\centering\cite{zhang-etal-2023-novel,ma-etal-2023-dreeam}} \\ 
T5 & - & 2020 & 2 & 10\% & \textbf{0.50} & \textbf{3.21\%} & \parbox{8cm}{\centering\cite{wadhwa-etal-2023-revisiting,najafi-fyshe-2023-weakly}} \\ 
GPT & 3,5 & 2022 & 2 & 10\% & \textbf{1.00} & \textbf{6.41\%} & \parbox{8cm}{\centering\cite{xu-etal-2023-s2ynre,wadhwa-etal-2023-revisiting}} \\ 
PubMedBERT & - & 2020 & 1 & 5\% & \textbf{0.25} & \textbf{1.60\%} & \parbox{8cm}{\centering\cite{zhou-etal-2023-continual}} \\ 
BioLinkBERT & large & 2022 & 1 & 5\% & \textbf{0.50} & \textbf{3.21\%} & \parbox{8cm}{\centering\cite{xu-etal-2023-nli}} \\ 
AnchiBERT & - & 2023 & 1 & 5\% & \textbf{1.00} & \textbf{6.41\%} & \parbox{8cm}{\centering\cite{yang-etal-2023-histred}} \\ 
mBART & 50 & 2020 & 1 & 5\% & \textbf{0.25} & \textbf{1.60\%} & \parbox{8cm}{\centering\cite{huguet-cabot-etal-2023-red}} \\ 
ViT & base & 2021 & 1 & 5\% & \textbf{0.33} & \textbf{2.12\%} & \parbox{8cm}{\centering\cite{wu-etal-2023-information}} \\ \hline
\end{tabular}
\end{adjustbox}
\end{table}

This underscores the prevalence of BERT-based methods being the most widespread in most literature. The preference for BERT-based models in the realm of RE can be attributed to their inherent characteristics that align seamlessly with the nature of RE systems:

\begin{itemize}
    \item BERT is pre-trained using two objectives: MLM and NSP. The MLM aspect predicts missing words in a sentence, while NSP helps the model understand relationships between consecutive sentences. This process aligns closely with the essence of relation extraction (Section \ref{sec:re}), wherein the objective is to identify a relation between two entities-a concept closely mirrored in BERT's training strategies.
    \item During MLM training, BERT randomly masks specific tokens in each sequence, enhancing the model's generalizability. This characteristic proves to be particularly beneficial for few-shot problems in RE.
    \item BERT uses unique tokens ([CLS] and [SEP]) to indicate the beginning and separation of sentences. Those tokens ease the delineation of head and tail entities in relation extraction. Many state-of-the-art results in RE are built upon adaptations of these tokens in the pretraining strategy.
\end{itemize}

Most papers on language adoption focus on English. However, there is also significant development in Asian languages. Notably, there are specific BERT models tailored for Chinese, such as ChineseBERT, and models like KLUE-RE and AnchiBERT, designed for Korean and Hanja, respectively, are emerging. 

Finally, it is interesting to note the ongoing presence of traditional word embedding techniques, such as GloVe and Word2Vec. However, these methods, along with CNN-based approaches, are primarily associated with the earlier years of the survey. Specifically, GloVe appears in four papers: two from 2020 and two from 2021, while Word2Vec is represented by only one paper from 2021. As for CNN-based approaches, as reflected in Table \ref{tab:summary}, there is a noticeable trend in their use during the early period of the survey. However, their usage has declined in recent years, indicating a reduced interest compared to BERT-based techniques in contemporary research.

\subsection{RQ3: Which datasets are used as benchmarks for RE using language models?}

Table \ref{tab:dataset-usage} presents the utilization of datasets in ACL conferences. Addressing RQ3 directly, the most prevalent benchmark datasets for RE have been TACRED, DocRed, and FewRel. Examining datasets with over ten recent cutting-edge RE research mentions reveals an intriguing pattern. 

Firstly, in traditional sentence-level RE, TACRED serves as the most robust benchmark. Researchers are actively exploring new techniques, including comparisons between joint models and pipeline approaches,  methods for sentence encoding, and advancements in accurately marking the tails and heads of relations for each model using this dataset.

Secondly, the domain of document-level relation extraction is gaining significant attention. DocRed, as a dataset, is instrumental in testing novel approaches and assessing recent advances. Researchers leverage this dataset to compare against baselines, fostering the ongoing evolution of methodologies in document-level RE.

\label{sec:rq3}
\begin{table}[H]
    \centering
    \caption{Dataset Usage in Papers}
    \label{tab:dataset-usage}
    \begin{tabular}{lcc}
        \toprule
        \textbf{Dataset} & \textbf{Times used} & \textbf{Papers that used the dataset} \\
        \midrule
        TACRED & 18 & \cite{zhao-etal-2023-improving,zhao-etal-2023-actively,xu-etal-2023-s2ynre,zhao-etal-2023-open,qin-joty-2022-continual,liu-etal-2022-pre,ma-etal-2021-sent,brody-etal-2021-towards,zhao-etal-2021-relation,eyal-etal-2021-bootstrapping,hu-etal-2021-semi-supervised,hu-etal-2021-gradient,yang-song-2022-fpc,zhou-chen-2022-improved,teru-2023-semi,wang-etal-2022-rely,liang-etal-2022-modeling,liu-etal-2022-hiure} \\
        DocRED & 15 & \cite{chen-etal-2023-models,zhang-etal-2023-novel,wadhwa-etal-2023-revisiting,sun-etal-2023-uncertainty,huang-etal-2021-three,huang-etal-2021-entity,nan-etal-2020-reasoning,eberts-ulges-2021-end,eyal-etal-2021-bootstrapping,ma-etal-2023-dreeam,guo-etal-2023-towards,xiao-etal-2022-sais,xu-etal-2022-document,xu-choi-2022-modeling,popovic-farber-2022-shot} \\
        FewRel & 12 & \cite{zhao-etal-2023-improving,zhao-etal-2023-actively,zhao-etal-2023-matching,zhao-etal-2023-open,qin-joty-2022-continual,liu-etal-2022-pre,yang-etal-2021-entity,brody-etal-2021-towards,zhao-etal-2021-relation,han-etal-2021-exploring,najafi-fyshe-2023-weakly,chen-li-2021-zs} \\
        SemEval 2010 - Task 8 & 10 & \cite{xu-etal-2023-s2ynre,zheng-etal-2023-jointprop,tian-etal-2021-dependency,yuan-eldardiry-2021-unsupervised,hu-etal-2021-semi-supervised,hu-etal-2021-gradient,yang-song-2022-fpc,teru-2023-semi,wang-etal-2022-rely,liang-etal-2022-modeling} \\
        ACE05 & 9 & \cite{zheng-etal-2023-jointprop,ye-etal-2022-packed,wang-etal-2021-unire,tian-etal-2021-dependency,pouran-ben-veyseh-etal-2020-exploiting,fu-grishman-2021-learning,yan-etal-2021-partition,yan-etal-2022-empirical,zhong-chen-2021-frustratingly} \\
        SciERC & 9 & \cite{zheng-etal-2023-jointprop,ye-etal-2022-packed,bassignana-plank-2022-mean,wang-etal-2021-unire,pouran-ben-veyseh-etal-2020-exploiting,yan-etal-2021-partition,yan-etal-2022-empirical,zhong-chen-2021-frustratingly,popovic-farber-2022-shot} \\
        NYT & 8 & \cite{wadhwa-etal-2023-revisiting,xie-etal-2021-revisiting,ma-etal-2021-sent,dai-etal-2021-two,yan-etal-2021-partition,hao-etal-2021-knowing,sui-etal-2021-distantly-supervised,christopoulou-etal-2021-distantly} \\
        Tacred-Revisited & 5 & \cite{xu-etal-2023-s2ynre,yang-song-2022-fpc,zhou-chen-2022-improved,wang-etal-2022-rely,liang-etal-2022-modeling} \\
        Re-DocRED & 4 & \cite{zhang-etal-2023-novel,zhou-etal-2023-continual,sun-etal-2023-uncertainty,guo-etal-2023-towards} \\
        ACE04 & 4 & \cite{ye-etal-2022-packed,wang-etal-2021-unire,yan-etal-2021-partition,zhong-chen-2021-frustratingly} \\
        Re-TACRED & 4 & \cite{xu-etal-2023-s2ynre,yang-song-2022-fpc,zhou-chen-2022-improved,wang-etal-2022-rely} \\
        MNRE & 3 & \cite{zheng-etal-2023-rethinking,wu-etal-2023-information,hu-etal-2023-multimodal} \\
        GDA & 3 & \cite{huang-etal-2021-three,nan-etal-2020-reasoning,xiao-etal-2022-sais} \\
        CDR & 3 & \cite{huang-etal-2021-three,nan-etal-2020-reasoning,xiao-etal-2022-sais} \\
        Wiki-ZSL & 3 & \cite{zhao-etal-2023-matching,najafi-fyshe-2023-weakly,chen-li-2021-zs} \\
        FB-NYT & 3 & \cite{shahbazi-etal-2020-relation,yuan-eldardiry-2021-unsupervised,liu-etal-2022-hiure} \\
        ChemProt & 2 & \cite{xu-etal-2023-nli,xu-etal-2023-s2ynre} \\
        ConLL & 2 & \cite{zheng-etal-2023-jointprop,wadhwa-etal-2023-revisiting} \\
        ADE & 2 & \cite{wadhwa-etal-2023-revisiting,yan-etal-2021-partition} \\
        DialogRE & 2 & \cite{yu-etal-2020-dialogue,xu-etal-2022-document} \\
        DDI & 1 & \cite{xu-etal-2023-nli} \\
        GAD & 1 & \cite{xu-etal-2023-nli} \\
        HistRED & 1 & \cite{yang-etal-2023-histred} \\
        MultiTacred & 1 & \cite{hennig-etal-2023-multitacred} \\
        SRED-FM & 1 & \cite{huguet-cabot-etal-2023-red} \\
        RED-FM & 1 & \cite{huguet-cabot-etal-2023-red} \\
        Twitter-2015 & 1 & \cite{zheng-etal-2023-rethinking} \\
        Twitter-2017 & 1 & \cite{zheng-etal-2023-rethinking} \\
        RISeC & 1 & \cite{gururaja-etal-2023-linguistic} \\
        EFGC & 1 & \cite{gururaja-etal-2023-linguistic} \\
        MSCorpus & 1 & \cite{gururaja-etal-2023-linguistic} \\
        Wiki80 & 1 & \cite{xu-etal-2023-s2ynre} \\
        BioRED & 1 & \cite{zhou-etal-2023-continual} \\
        MATRES & 1 & \cite{yuan-etal-2023-discriminative} \\
        EventStoryLine & 1 & \cite{yuan-etal-2023-discriminative} \\
        Causal-TimeBank & 1 & \cite{yuan-etal-2023-discriminative} \\
        MAWPS & 1 & \cite{jie-etal-2022-learning} \\
        Math23k & 1 & \cite{jie-etal-2022-learning} \\
        MathQA & 1 & \cite{jie-etal-2022-learning} \\
        SVAMP & 1 & \cite{jie-etal-2022-learning} \\
        SemEval 2018 Task 7 & 1 & \cite{bassignana-plank-2022-mean} \\
        SKE & 1 & \cite{xie-etal-2021-revisiting} \\
        FOBIE & 1 & \cite{kruiper-etal-2020-laymans} \\
        SF & 1 & \cite{yu-etal-2020-dialogue} \\
        SPOUSE & 1 & \cite{pouran-ben-veyseh-etal-2020-exploiting} \\
        ERE & 1 & \cite{fu-grishman-2021-learning} \\
        FUNDS & 1 & \cite{zhang-etal-2021-entity} \\
        GIDS & 1 & \cite{hao-etal-2021-knowing} \\
        SMiLER & 1 & \cite{seganti-etal-2021-multilingual} \\
        RE-QA & 1 & \cite{najafi-fyshe-2023-weakly} \\
        KBP37 & 1 & \cite{teru-2023-semi} \\
        WikiDistant & 1 & \cite{christopoulou-etal-2021-distantly} \\
        DrugCombination & 1 & \cite{tiktinsky-etal-2022-dataset} \\
        DWIE & 1 & \cite{xu-choi-2022-modeling} \\
        \bottomrule
    \end{tabular}
    \caption{Most widespread benchmarks datasets}
\end{table}

Lastly, the FewRel dataset has emerged as a benchmark for few-shot relation extraction. This dataset plays a crucial role in evaluating the capabilities of both standard and large language models. The discussions surrounding the effectiveness of LMS and the foray of LLMs in capturing previously unseen relations underscore the dataset's importance in shaping the discourse on few-shot relation extraction.

Exploring the long tail of datasets with only a single mention in cutting-edge relation extraction reveals a diverse landscape. These datasets, spanning diverse domains such as mathematics, dialogues, natural disasters, drugs, and heritage, signal the broad interest in employing relation extraction techniques across various academic and societal domains. The presence of datasets in these unique and specialized areas is compelling evidence of the technique's applicability and relevance to a wide array of fields within academia and society.

\begin{table}[H]
    \centering
    \caption{Statistics and a summary of the most widely used datasets for Relation Extraction}
    \small
    \begin{adjustbox}{max width=\textwidth}
    \begin{tabular}{p{1.8cm}p{2.5cm}p{2.5cm}cccp{2.2cm}p{2.2cm}p{2.2cm}p{2.2cm}}
        \toprule
        Dataset & RE Task & Dataset based on & \# of relations & Example relations & Relation mentions & Corpus size & Train & Dev & Test \\
        \midrule
        TACRED & Multipurpose - Sentence Level RE & Newswire and web text manually annotated using Amazon Mechanical Turk crowdsourcing. Covers common relations between people, organizations, and locations. & 43 & per:schools\_attended and org:members & 21,784 & 106,264 & 68,124 & 22,631 & 15,509 \\
        \addlinespace
        DocRED & Document-Level RE & Wikipedia data, both manually and distant supervised annotated. & 96 & educated\_at, spouse, creator, publication\_date... & 155,535 & Manually labelled: 5,053 Distant supervised: 101,873 & Manually labeled: 3,053 Distant supervised: 101,873 & 1,000 & 1,000 \\
        \addlinespace
        FewRel & Few Shot RE & 70,000 sentences on relations derived from Wikipedia and annotated by crowdworkers. & 100 & member\_of, capital\_of, birth\_name & 58,267 & 70,000 & 44,800 & 6,400 & 14,000 \\
        \addlinespace
        SemEval 2010 - Task 8 & Multipurpose - Sentence Level RE & Semantic relations between pairs of nominals of general purpose. & 9 & Entity-Origin, entity-destination, cause-effect... & 6,674 & 10,717 & 8,000 & - & 2,717 \\
        \addlinespace
        ACE05 &  Multilingual - Sentence Level RE &  1,800 files encompassing diverse English, Arabic, and Chinese genres. These texts have been meticulously annotated for entities, relations, and events. This extensive collection is the entirety of the training data utilized during the 2005 Automatic Content Extraction (ACE) technology evaluation for these languages. & 18 & Person-Social , Organization-Affiliation, Agent-Artifact & 7,105  &  10,573  &7,273   & 1,765 & 1,535 \\
        \addlinespace
        SciERC & Multipurpose - Low Resource & Collection of 500 scientific abstracts annotated in terms of entities and relations & 7 & usef\_for , hyponym\_of, feature\_of, conjunction & 4,716 & 500 & - & - & - \\
        \addlinespace
        NYT-FB & Multipurpose - Distant Supervision RE & New York Times news annotated by distant supervision. & 52 & nationality, place\_of\_birth & 142,823 & 717,219 & 455,771 & - & 172,448 \\
        \bottomrule
    \end{tabular}
    \end{adjustbox}
    \label{tab:relation-datasets-statsitcs}
\end{table}

We conducted a thorough analysis, focusing on datasets with a minimum of five references in the literature, to provide an in-depth exploration of the landscape of RE (Table \ref{tab:relation-datasets-statsitcs}). Our systematic examination covers various dimensions, including the number and types of relations and relevant statistical metrics. We meticulously consider factors such as the source of information, domain specificity, nature of relations, and other pertinent attributes for each dataset. By delving into the intricacies of these datasets, our goal is to offer a comprehensive overview that illuminates their diverse characteristics, providing valuable insights and contributing to a nuanced understanding of the field of RE.

\subsection{RQ4: Are new large language models like GPT or T5 useful in RE versus widespread models such as BERT?}

To address RQ4, we analyzed the performance of various models on three benchmark datasets in Section \ref{sec:llmvslm} and reviewed the most widespread models used in Table \ref{tab:model-usage}. The results shed light on the effectiveness of new large language models like GPT and T5 compared to well-established models like BERT.

Despite their promising capabilities, large language models, including GPT and T5, are used less frequently than models such as RoBERTa and BERT. This phenomenon could be attributed to the inherent suitability of these language models for downstream tasks like chatbots, where their remarkable ability to capture extensive contextual information is more prominently leveraged. The current landscape suggests that large language models do not play a central role in advancing state-of-the-art performance in extraction tasks. However, it is essential to note that the field is dynamic, and the influence of these models may evolve with ongoing research and advancements.

An intriguing observation arises when comparing the insights from Section \ref{sec:llmvslm} with the utilization trends outlined in Table \ref{tab:model-usage}. Notably, state-of-the-art results are predominantly achieved by models based on RoBERTa. Surprisingly, RoBERTa, despite its demonstrated strength in terms of results, is underused in contrast to the widespread adoption of BERT. This discrepancy in adoption could be due to BERT's adaptability, which encourages a more profound exploration of diverse approaches for encoding relations and comparison against other baseline models. In contrast, RoBERTa's superior performance might result in fewer variations in its application, as it already delivers robust results without requiring extensive modifications.

The landscape of relation extraction is currently marked by a nuanced interplay between language models' performance and practical adoption. While large language models show promise, their utilization in cutting-edge relation extraction remains modest. The interplay of factors such as adaptability, performance, and the specific demands of downstream tasks shapes the landscape, suggesting a need for ongoing exploration and fine-tuning of these models about extraction.

To ensure this analysis is as up-to-date as possible, we used the advanced search feature of Web of Science to perform four different queries. Each query included "relation extraction" in the title and searched for specific models and their variants in the abstract or topic. For example, the query for T5 was: \textit{TI=("relation extraction") AND (TS=(T5 OR "Text-To-Text Transfer Transformer") OR AB=(T5 OR "Text-To-Text Transfer Transformer"))}. For BERT, we found 31 papers published in the last two years, 2023 and 2024. In contrast, we found only three papers each for T5 and GPT. These results support more robust conclusions regarding the predominance of BERT in our survey findings.

\subsection{Synthesis}

We have identified two primary methods of RE: joint and pipeline. Joint models are designed to simultaneously identify entities and relationships, capturing the inherent interdependence between these two tasks. In contrast, the pipeline approach follows a sequential process, starting with NER to identify entities, followed by at least a classification stage for these recognized entities. The pipeline approach can involve up to three stages, with an additional step of relation identification between NER and relation classification, as depicted in Table \ref{tab:model-usage-acl} and Table \ref{tab:model-usage-others}. While this two-step or three-step process allows for more specialized handling of each task, it also introduces the risk of error accumulation. In recent years, joint approaches have gained more attention, mainly due to concerns that the pipeline method is prone to error propagation.

Future trends in the realm of new language models, including sub-models and large language models, point toward increasingly complex and intricate models with more parameters. For instance, LLaMA 3 has 70 billion parameters \cite{dubey2024llama}, GPT-4 has 1.8 trillion parameters \cite{openai2303gpt}, and Gemini supports a context window of 128,000 tokens with billions of parameters \cite{team2023gemini}. These models will seem small compared to future developments in our rapidly evolving space. In the context of RE, these advanced models, with their extended context windows, will be particularly valuable for document-level tasks. They can maintain entire documents within their context, improving the handling of extensive information. Furthermore, in multilingual and multimodal settings, the vast context capabilities of these LLMs will enhance understanding across different languages and modalities. Additionally, these models can offer significant advantages for few-shot or low-resource scenarios by augmenting evidence for underrepresented domains and improving generalization with minimal data.

Before concluding this section, it is essential to address this research's ethical considerations and global impact. All reviewed papers adhere to ethical standards by utilizing anonymized or publicly available data, ensuring their contributions to society are made responsibly. They are committed to minimizing bias and avoiding potential ethical issues, reflecting a conscientious approach to research and its broader implications. Finally, the environmental impact is minimal, considering the CO2 emissions and energy consumption associated with the research discussed. While the reviewed papers do not explicitly provide this information, it is reasonable to infer that only foundational work and the initial training of large language models have a significant environmental footprint. Most of the studies leverage pre-trained models, thereby avoiding the resource-intensive process of training from scratch, which is the most demanding stage in terms of energy consumption.

\section{Conclusion}
\label{sec:conclusion}
This paper  reviews cutting-edge relation extraction, explicitly focusing on models leveraging language models. Our analysis encompasses 137 papers, spanning dataset proposals, novel techniques, and excluded studies. 

Our research focuses on 65 papers, which we have rigorously categorized into eight sub-tasks reflecting the latest advances in RE. From these papers, we have extracted crucial insights into training strategies, novel methodologies, and the utilization of language models, which have emerged as a cornerstone of contemporary RE research. 

Our findings highlight the dominance of BERT-based approaches to extraction, underscoring their efficacy in achieving state-of-the-art results. Notably, BERT has emerged as the most widely utilized model for extraction to date. Additionally, emerging large language models like T5 exhibit promise, particularly in the context of few-shot relation extraction, showcasing their ability to capture previously unseen relations. However, these models are primarily suited for tasks such as text generation and question answering; their capability to handle large context windows suggests a promising future for developing RE techniques based on these models.

Language models are highly suitable for RE because they effectively capture the semantic nuances and complex relationships within text. By leveraging these models, research has been able to address the inherent challenges of RE, such as ambiguity and context sensitivity, providing robust solutions that enhance the accuracy and efficiency of extracting relations across diverse domains.

Throughout our exploration, we have identified FewRel, DocRed, and TACRED as pivotal benchmark datasets, aligning with the three primary themes driving contemporary research about extraction. FewRel serves as a benchmark for few-shot relation extraction, DocRed facilitates advancements in document-level relation extraction, and TACRED remains a stalwart in traditional sentence-level relation extraction. Collectively, these datasets offer a comprehensive evaluation ground for models addressing the diverse challenges in cutting-edge relation extraction.

Finally, although this is more related to the NER domain, further exploration of issues related to the number of entities in different domains is needed. We recognize that a relation's domain size can significantly impact model performance. For example, relations with large domains, such as ``child of'' or ``contains,'' require models good at generalizing observations. On the other hand, relations with smaller domains, like ``share a common border,'' might benefit from models with sufficient memory capacity. This observation indicates that a one-size-fits-all approach might not be the best, and adjusting models to fit the specific characteristics of relation domains could improve performance and generalization.

\section*{Acknowledgment}

The research reported in this paper was supported by the DesinfoScan project: Grant TED2021-129402B-C21 funded by  MCIN /AEI/ 10.13039/ 501100011033 and, by the European Union \\ NextGenerationEU/PRTR, and FederaMed project: Grant PID2021-123960OB-I00 funded by MCIN/ AEI/ 10.13039/501100011033 and by ERDF A way of making Europe. Finally, the research reported in this paper is also funded by the European Union (BAG-INTEL project, grant agreement no. 101121309).

%
%
%
\bibliographystyle{splncs04}
\bibliography{bibimpu}

\end{document}